\documentclass[10pt,journal,compsoc]{IEEEtran}

\usepackage{amssymb,times}
\usepackage{graphicx}
\usepackage{color}
\usepackage[cmex10]{amsmath}

\usepackage{algorithm}
\usepackage{algpseudocode}

\ifCLASSOPTIONcompsoc
  \usepackage[nocompress]{cite}
\else
  \usepackage{cite}
\fi

\ifCLASSOPTIONcompsoc
  \usepackage[caption=false,font=footnotesize,labelfont=sf,textfont=sf]{subfig}
\else
  \usepackage[caption=false,font=footnotesize]{subfig}
\fi

\DeclareMathOperator{\vect}{vec}

\begin{document}
\title{DART: Distribution Aware Retinal Transform for Event-based Cameras}

\author{Bharath~Ramesh*, Hong~Yang, Garrick~Orchard, Ngoc~Anh~Le~Thi, Shihao~Zhang,
        and~Cheng~Xiang,~\IEEEmembership{Member,~IEEE}
\IEEEcompsocitemizethanks{\IEEEcompsocthanksitem B. Ramesh, Hong Yang, and Garrick Orchard, are with Temasek Laboratories,
National University of Singapore. Le Thi Ngoc Anh, Shihao Zhang and Cheng Xiang are with the Department of Electrical and Computer Engineering, National University of Singapore.\protect\\
*E-mail: bharath.ramesh03@u.nus.edu (corresponding author)
}
\thanks{Manuscript received }}

\markboth{IEEE Transactions on Pattern Analysis and Machine Intelligence}%
{Ramesh \MakeLowercase{\textit{et al.}}: DART: Distribution Aware Retinal Transform for Event-based Cameras}

\IEEEtitleabstractindextext{%
\begin{abstract}
We introduce a generic visual descriptor, termed as distribution aware retinal transform (DART), that encodes the structural context using log-polar grids for event cameras. The DART descriptor is applied to four different problems, namely object classification, tracking, detection and feature matching: (1) The DART features are directly employed as local descriptors in a bag-of-features classification framework and testing is carried out on four standard event-based object datasets (N-MNIST, MNIST-DVS, CIFAR10-DVS, NCaltech-101). (2) Extending the classification system, tracking is demonstrated using two key novelties: (i) For overcoming the low-sample problem for the one-shot learning of a binary classifier, statistical bootstrapping is leveraged with online learning; (ii) To achieve tracker robustness, the scale and rotation equivariance property of the DART descriptors is exploited for the one-shot learning. (3) To solve the long-term object tracking problem, an object detector is designed using the principle of cluster majority voting. The detection scheme is then combined with the tracker to result in a high intersection-over-union score with augmented ground truth annotations on the publicly available event camera dataset. (4) Finally, the event context encoded by DART greatly simplifies the feature correspondence problem, especially for spatio-temporal slices far apart in time, which has not been explicitly tackled in the event-based vision domain. 
\end{abstract}

}

\maketitle

\IEEEdisplaynontitleabstractindextext

\IEEEpeerreviewmaketitle

\IEEEraisesectionheading{\section{Introduction}\label{sec:introduction}}

\IEEEPARstart{O}{bject} classification and tracking are important problems in machine vision with applications ranging from surveillance, human computer interaction, to medical imaging. These two interrelated problems receive a lot of attention from the research community. In particular, object classification can be treated as a sub-problem within object tracking, when discriminative models are used to track objects over time. Given the initial state (e.g., position and extent) of a target object in the first frame, the goal of tracking is to estimate the states of the target in the subsequent frames. Numerous factors affect the performance of a tracking algorithm, including view-point variation, occlusion, as well as background clutter. Although these issues have been studied for several decades, real-time processing with conventional frame-based video cameras that acquire largely redundant data at high sampling rates remains difficult without dedicated hardware. For instance, Ren~\textit{et~al.} \cite{Ren2017}, a state-of-the-art object detection method on PASCAL VOC 2007, 2010 and MS COCO datasets, runs only at 5 fps on an NVIDIA Tesla K40 GPU.
\par
Silicon retinas or event cameras, such as the Asynchronous Time-based Image Sensor (ATIS) \cite{Posch2008}, are fundamentally different from traditional cameras that output a sequence of frames at fixed intervals. The term `event' refers to a spike output that is characterized by a spatial location ($x,y$), timestamp ($t$) and polarity of the brightness change ($p$). Thus, the output of an event camera is a stream of asynchronous spikes that are triggered by brightness changes sensed by individual pixels. Naturally, events are most likely to occur at the edges that delineate the structures in the scene, and to recognize individual objects amidst noisy events is a challenging problem.
\par
Object recognition has been a central task to the vision community since the early days of using computers to identify hand-written characters \cite{Roberts1960}. Hence, object recognition research in its budding years was primarily concerned with 3D shape representation (e.g. \cite{Crowley1984}). Subsequent two decades of research in object recognition moved away from 3D geometry to appearance-based recognition systems, which opened up new horizons in recognizing natural images \cite{Andreopoulos2013}. This time-tested technique of using visual descriptors to recognize objects is a natural choice \cite{Andreopoulos2013}, given an efficient and effective descriptor can be designed for asynchronous silicon retinas. 
\par
\begin{figure}[t]%
\centering
\includegraphics[width=3in]{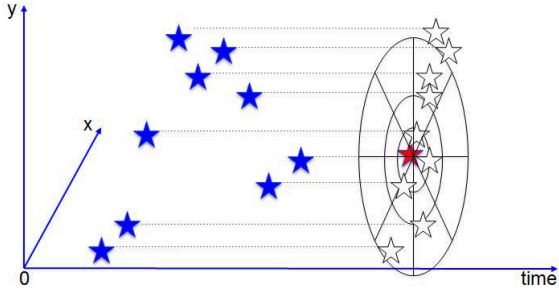}%
\caption{The most recent event is indicated in red and the previous events are indicated in blue. White stars indicate the position of previous events mapped onto the log-polar bins at the current time.}%
\label{fig:primate}%
\end{figure}

This paper present an event-based structural descriptor using a log-polar grid, termed as distribution aware retinal transform (DART), which simulates the distribution of cones in the primate fovea \cite{Schwartz1977}. As shown in Fig. \ref{fig:primate}, a log-polar grid is centered at the latest event and the past events, whose space-time coordinates are marked as a `star', are binned into nearest spatial locations of the grid. Subsequently, the DART descriptor is formed using the overall interpolated event count within each bin of the log-polar grid. As the neuromorphic camera responds to changes in log-intensity, a brighter contrast or a faster motion results in an increased event rate. Thus, normalization of the DART descriptor is critical to capture the relative distribution of the surrounding events, and to account for camera and object motion, the descriptor is updated on an event-by-event basis using a queue to capture precise space-time information.
\par
To the best of our knowledge, this paper represents a significant step forward in computing a structural descriptor for event-based data. Similar to \cite{Lagorce2016}, each event now is a quintuple (space-time coordinates, polarity and descriptor) that can be used to recognize a set of events. The contributions of this paper are as follows:
\begin{itemize}
	\item A log-polar based feature descriptor that is robust to scale, rotation and view-point variations is presented for event cameras. The DART-driven approach reports the best classification compared to existing works on the MNIST-DVS (99\%), CIFAR10-DVS (65.43\%) and NCaltech-101 (70.33\%) datasets. Moreover, real-time unconstrained view-point object classification is demonstrated for silicon retinas as a first.
	\item Using the DART descriptor, an event-based long-term object tracking (eLOT) framework, consisting of a local search \emph{tracker} and a global search \emph{detector}, is proposed with online learning to account for appearance changes over time. The eLOT system is evaluated on the event camera dataset \cite{Mueggler2017} with augmented ground truth annotations for various camera motion profiles such as translation, rotation, 6-DOF, and thus is one of the first tracking benchmarks for the research community. 
	\begin{itemize}
		\item The \emph{tracker} uses one-shot learning with statistical bootstrapping of circular shifted DART descriptors to obtain a robust object representation. 
		\item The \emph{detector} outputs a candidate object location with maximum spatial confidence for rescuing tracker fails. 
	\end{itemize} 
	\item Feature matching is demonstrated for event cameras with potential applications to many vision problems. With a high temporal resolution, matching events can be considered as a local update or a model search problem, but matching far apart time slices is non-trivial and our results pave way to solving intricate problems like recognizing previously encountered scenes. 
\end{itemize}
\par
The rest of the paper is organized as follows. Sec.~\ref{sec:2} describes the event cameras and the related work. Sec.~\ref{sec:3} outlines the DART descriptor computation in detail, followed by its application to object classification in Sec.~\ref{sec:4}. Next, Sec.~\ref{sec:5} explains the eLOT system consisting of the tracker and the detector modules. Then in Sec.~\ref{sec:6}, DART-based feature matching for event cameras is briefly discussed. Sec.~\ref{sec:7} and Sec.~\ref{sec:8} reports the experimental setup and results respectively. Finally, the paper is concluded in Sec.~\ref{sec:9}.

\section{Event Cameras}
\label{sec:2}
For real-time experiments, we use the commercial event camera, the Dynamic and Active-pixel Vision Sensor (DAVIS) \cite{Brandli2014}. It has 240 $\times$ 180 resolution, 130 dB dynamic range and 3 microsecond latency. The DAVIS can concurrently output a stream of events and frame-based intensity read-outs using the same pixel array. An event consists of a pixel location ($x$, $y$), a binary polarity value ($p$) for positive or negative change in log intensity and a timestamp in microseconds ($t$). In this work, polarity of the events are not considered, and only the event stream of the DAVIS is used.
\subsection{Related Work}
\textbf{Feature Extraction.} By simulating the non-uniform distribution of cones in the primate fovea, log-polar grids arguably offer a way of encoding similar to that of the human vision system \cite{Braccini1982}. Log-polar transform was introduced in the image processing domain by \cite{Weiman1979197} and further rigorously formulated by \cite{Messner1985}. Subsequently, grayscale sampling using log-polar imaging has been applied to various problems with moderate success: object recognition \cite{Tistarelli1995,Tistarelli1997,bone2006position,Lin2015}, robotic vision \cite{Sandini1980,Grosso2000}, etc., due to the attractive property of scale and rotation invariance when segmentation is available \cite{Ramesh2015}. Nonetheless, event cameras do not output grayscale frames for direct application of log-polar transform. In order to obtain a descriptor robust to scale, rotation, and view-point variations, we propose event-based spatio-temporal log-polar histograms (Fig. \ref{fig:primate}).
\par
Log-polar histograms have been successfully applied as local descriptors, such as shape context \cite{shapecontext_belongie} and self-similarity descriptor \cite{selfsimilarity_shechtman}. The closely related work of shape context \cite{shapecontext_belongie} creates log-polar histograms on binary images. Invariance to translation is intrinsic to local log-polar grids since all measurements are taken with respect to a single point or an event, whereas to achieve scale invariance in shape context, all radial distances between the point pairs in the shape have to be normalized by the mean distance. In contrast, DART considers a fixed log-polar grid to obtain the descriptors and therefore it is computationally easy to achieve with a look-up table. Moreover, shape context uses a relative frame, based on treating the tangent vector at each point as the positive x-axis to achieve rotation invariance. On the contrary, the DART descriptors are designed only to be equivariant, i.e., the log-polar histograms of two events with different scale/rotations result in cyclical shifts in the log-polar domain.
\par
\textbf{Object Classification.} In the neuromorphic community, the use of descriptors for object classification is gaining momentum. Examples are time-surfaces \cite{Lagorce2016}, a time oriented approach to extract spatio-temporal features that are dependent on the direction and speed of motion of the objects; ripple pond networks \cite{Afshar2013} that perform a transformation converting two dimensional images to one dimensional temporal patterns. The major drawback of these works is either the dependence of feature extraction on motion in the case of time-surfaces, or the need for precise centering of a salient object in the case of the ripple pond network. In particular, a faster/slower motion leads to a different data rate and the time-constants of the exponential kernel used by the time surfaces needs to be changed accordingly, as acknowledged by its authors. In this work, we avoid both these issues by centering the log-polar grid on an incoming event and letting the number of events be the deciding factor for feature extraction instead of choosing a time interval or a decay rate. 
\par
A recent event-based \cite{Zou2017} work describes creating and matching event streams across a stereo event-camera pair and shows good performance. However, the running time for creating a single event map for matching is about 25 ms on a PC with Intel Core i7 Quad processor (3.40GHz, 4 cores) with 20 GB memory. In contrast, we show real-time performance of our DART method on an Intel Compute Stick, which uses an Intel Core m5-6Y57 vPro processor with 2 GB memory, by matching events across a single event camera. Lastly, spatio-temporal feature extraction using echo-state networks \cite{Lagorce2015a} is a closely related in the sense that it works on event-based sparse and asynchronous input streams. This proof-of-concept study if implemented on configurable neuromorphic platforms with online learning capabilities can be used in real-world event-based vision applications, such as recognition of objects and sequences, but remains a topic of ongoing research. 
\par
\textbf{Object Tracking.} Object tracking with standard cameras leads to a commonly encountered dilemma: lower frame rates give rise to imprecise/failed tracking due to large relative object movement between successive images, or on the contrary, higher frame rates burdens real-time requirements. Thus, event-based tracking solves both these issue with asynchronous pixels with very high temporal resolution in the order of microseconds. Object tracking works using event-based data are part-based \cite{Valeiras2015} or kernel-based \cite{Lagorce2015} methods that track incoming blobs of events based on local shape properties. Recently, feature or corner tracking has been addressed to some extent \cite{Zhu2017,Mueggler2017a}. However, tracking of objects with a moving event camera with arbitrary motion profile has remained an untackled problem. Taking a step in this direction, the aim of this paper is to track specific patterns/objects as a whole undergoing arbitrary 6-DOF motion. In particular, the tracking problem is posed as a local update classification task, inspired by how discriminative trackers \cite{Henriques2015} update the object position over time. The lack of a tracking dataset with ground truth annotations is also recognized by us and we release full-length ground truth annotations for the shapes data in the event camera dataset \cite{Mueggler2017} and create an important benchmark for the research community. 
\par
\textbf{Object Detection.} As event-based vision is relatively new, only a limited amount of work addresses object detection, which is critical for long-term object tracking for handling track failures and object occlusion, using these devices \cite{Liu2016,Lenz2018}. Liu \emph{et al}. \cite{Liu2016} focuses on combining a frame-based CNN detector to facilitate the event-based module. We argue that using deep neural networks for event-based object detection may achieve good performance with lots of training data and computing power, but they go against the idea of low-latency, low-power event-based vision. In contrast, \cite{Lenz2018} presents a practical event-based approach to face detection by looking for pairs of blinking eyes. While \cite{Lenz2018} is applicable to human faces in the presence of activity, we develop a general purpose event-based, object detection method to deal with the case of re-tracking a lost or occluded object. Therefore, similar in spirit to the seminal work on tracking, learning and detection for frame-based cameras \cite{Kalal2012}, the proposed object tracking framework is a complete long-term tracking solution.
\section{DART Descriptor}
\label{sec:3}
Each incoming event, ${\bf{e}}_i = (x_i,y_i,t_i,p_i)^T$ with pixel location $x_i$ and $y_i$, timestamp $t_i$, polarity $p_i$, is encoded as a feature vector ${\bf{x}}_i$. In this work, polarity of the events are not considered. To deal with hardware-level noise from the event camera, two filtering steps are used: (1) nearest neighbour filtering and (2) refractory filtering. We define a spatial Euclidean distance between events as, 

\begin{equation}
{D_{i,j}} = \left| {\left| {\left( {\begin{array}{*{20}{c}}
{{x_i}}\\
{{y_i}}
\end{array}} \right) - \left( {\begin{array}{*{20}{c}}
{{x_j}}\\
{{y_j}}
\end{array}} \right)} \right|} \right| ~.
\label{eq:distanceevents}
\end{equation}
Using the above distance measure, for any event a set of previous events within a spatial neighborhood can be defined as, $N\left( {{\bf{e}}_i,\gamma } \right) = {\rm{\{ }}{\bf{e}}_i\;{\rm{|}}\;j < i,\;{D_{i,j}} < \gamma \} {\rm{\;}}$, where $\gamma = \sqrt{2}$ for an eight-connected pixel neighbourhood. When the time difference between the current event and the most recent neighboring event is less than a threshold, ${{\bf{\Theta }}_{noise}}$, the filter can be written as
\begin{equation}
{F_{noise}}\left( {\bf{e}} \right) = \{ {{\bf{e}}_i}{\rm{|}}\;N({{\bf{e}}_i},\;\sqrt 2 )\backslash N({{\bf{e}}_i},\;0)\; \ni \;{{\bf{e}}_j}\;\} ~,
\label{eq:noisefilter}
\end{equation}
where ${t_i} - {t_j} < {{\bf{\Theta }}_{noise}}$. When the neighborhood is only the current pixel, $\gamma = 0$, the set of events getting through the refractory filter ${F_{ref}}$ are those such that,
\begin{equation}
{F_{ref}}\left( {\bf{e}} \right) = {\rm{\{ }}{\bf{e}}_j{\rm{|}}\;\;{t_i} - {t_j} > {{\rm{\Theta }}_{ref}}\;\forall \;j\;|\;{\bf{e}}_j \in N\left( {{\bf{e}}_j,\;0} \right)\}~. 
\label{eq:reffilter}
\end{equation}
Cascading the filters, we can write the filtered incoming events as,
\begin{equation}
\left\{ \hat{\bf{e}} \right\} = \;{F_{noise}}\left( {\;{F_{ref}}\left( {\bf{e}} \right)\;} \right) ~.
\label{eq:filtered}
\end{equation}
\par
To extract the DART features efficiently, a first-in, first-out (FIFO) queue structure is used to contain the event locations as they arrive while a count matrix is updated at these locations. Once the container is full, the queue is popped and the latest incoming event is pushed. We denote the count matrix as $\textbf{C} \in \mathrm{R}^{N \times M}$, in which each entry is updated as follows.
\begin{equation}
 {\begin{array}{*{20}{c}}
{{\rm{queue}}{\rm{.push(}}x,y{\rm{) \to ~~~~~~~~~~}}C(y,x) = C(y,x) + 1}\\
{{\rm{queue}}{\rm{.pop(}}{x_{old}}{\rm{,}}{y_{old}}{\rm{)\to}~}C({y_{old}},{x_{old}}) = C({y_{old}},{x_{old}}) - 1}
\end{array}}
\label{eq:queuematch}
\end{equation}
Let us define the mapping from Cartesian coordinates of the image - $C(y,x)$ to log-polar coordinates - $X(\rho,\theta)$ as follows, 
\begin{equation}
x'=r \cos\theta, \qquad y'=r\sin\theta  ,
\label{eqn:lptparam1}
\end{equation}
where $(r,\theta)$ are polar coordinates defined with $(x_c,y_c)$ as the center of the transform and $(x',y') = (x-x_c,y-y_c)$, that is, 
\begin{equation}
r= \sqrt{(x')^{2} + (y')^{2}}.
\label{eqn:lptparam2}
\end{equation}
\par
Let the radii of the smallest and the largest ring be represented as $r_{min}$ and $r_{max}$ respectively, and the number of rings and wedges chosen for the log-polar grid be $n_r$ and $n_w$ respectively. The logarithmic scaling is defined as $\rho= \log r$, given in terms of the $r_{min}$ and $r_{max}$ as
\begin{equation}
\rho_q = \exp{\left(\frac{q \times \log\left(\frac{r_{max}}{r_{min}}\right)}{n_r - 1}\right)} \times r_{min}~,
\label{eq:ringnumber}
\end{equation}
where $q = \{1,2, \cdots, n_r\}$ gives the distance of each ring to the center and by definition $\rho_0 = 0$. 
\par
The angle $\theta$ is required to be in the range $[0,2\pi)$, but $\arctan$ is defined only for $(\frac{-\pi}{2},\frac{\pi}{2})$. Therefore, the angles are computed depending on the quadrant to produce output in the range $(\frac{-\pi}{2},\frac{3\pi}{2}]$, which can be mapped to $[0,2\pi)$ by adding $2\pi$ to negative values. The angular step for each wedge is simply given by $\theta_{step} = 2\pi/n_w$. 
\par
Let each bin of the log-polar grid be represented by its mid-point $(\rho_q^m,\theta_p^m)$, which is obtained as $\rho_q^m = (\rho_{q-1}^m + \rho_q^m)/2$ and $\theta_p^m = (p\theta_{step} + (p-1)\theta_{step})/2$ where $p = \{1,2, \cdots, n_w\}$. Since the log-polar grid assumes sub-pixel locations on the image plane, mid-points can be spatially very close to each other around the center and thus a past event contributes to the four nearest bins around it. On the contrary, events occur only at integer locations due to the sensor design and for a given log-polar lattice design, nearest neighbour weights are calculated offline and a look-up table is used to distribute an event to its four nearest log-polar bins. 
\par
For an event $(\rho_i,\theta_i)$ with four neighboring log-polar bins, denoted as $(\rho_q^m,\theta_p^m)$, $(\rho_{q'}^m,\theta_p^m)$, $(\rho_q^m,\theta_{p'}^m)$, and $(\rho_{q'}^m,\theta_{p'}^m)$ where $p' = p +1$ and $q' = q + 1$, the weight of each log-polar bin is calculated using bi-linear interpolation coefficients as follows,
\begin{equation}
\begin{bmatrix}
b_{qp}\\b_{q'p}\\b_{qp'}\\b_{q'p'}
\end{bmatrix}=
\left(\begin{bmatrix}
1 & \rho_q^m & \theta_p^m & \rho_q^m \theta_p^m \\
1 & \rho_{q'}^m & \theta_p^m & \rho_{q'}^m \theta_p^m \\
1 & \rho_q^m & \theta_{p'}^m & \rho_q^m \theta_{p'}^m \\
1 & \rho_{q'}^m & \theta_{p'}^m & \rho_{q'}^m \theta_{p'}^m
\end{bmatrix}^{-1}\right)^{\rm T} \begin{bmatrix}
1\\\rho_i\\\theta_i\\\rho_i\theta_i
\end{bmatrix}.
\label{eq:bilinear}
\end{equation}
\begin{figure}[t]%
\centering
\includegraphics[width=3in]{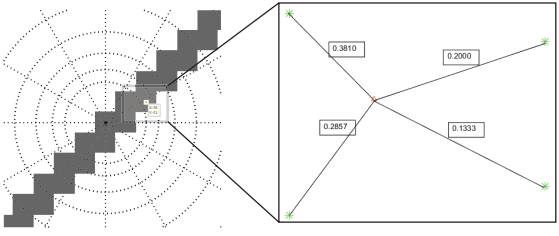}%
\caption{A log-polar grid placed at an incoming event occurring at a moving edge where gradient change is dominant. To obtain a robust feature representation, a past event (red point) is split into adjacent bins processing based on its distance to the bin centers (green asterisk).}%
\label{fig:dart_interp}%
\end{figure}
Note that $b_{qp}+b_{q'p}+b_{qp'}+b_{q'p'} = 1$ and in boundary cases where an event is not enclosed by four mid-points,  the closest mid-point weight $b_{qp}$ is set to one and the others are set to zero. Fig.~\ref{fig:dart_interp} illustrates this process with an example. 
\par
Each past event in the FIFO within the maximum radius of the log-polar grid centered at the latest event thus updates the DART representation $X_i \in \mathrm{R}^{n_r \times n_w}$,
\begin{equation}
\begin{split}
X_i(q,p) & =  X_i(q,p)  +   b_{qp}\\
X_i(q',p) & =  X_i(q',p)  +   b_{q'p}\\
X_i(q,p') & =  X_i(q,p')  +   b_{qp'}\\
X_i(q',p') & =  X_i(q',p')  +   b_{q'p'}
\end{split}
\label{eq:dartupdate}
\end{equation}
The $\ell_1$-normalization gives the feature vector,
\begin{equation}
{\bf{x}}_i = \vect(X_i) ~{\oslash} \left({\sum\limits_{q = 1}^{{n_r}}\sum\limits_{p = 1}^{{n_w}} {X_i(q,p)} }\right)
\label{eq:dart_norm}
\end{equation}
where ${\oslash}$ is the element-wise Hadamard division. The dimension of the feature descriptor is based on choice of the log-polar lattice, the number of rings and wedges. Nonetheless, the more crucial design parameter is the size of the FIFO that determines the extent to which motion affects the descriptor itself. In other words, the upper limit for the FIFO size is a quantity dependent on motion and is difficult to choose. Thus, selecting a lower limit is easier and we set it empirically by considering the number of rings and wedges of the log-polar grid. A typical choice for a 10 by 12 log-polar grid having 120 bins can be some fraction of the total number of bins, i.e., $S = 120 \times (\alpha \times 120)$, where $\alpha$ ranges from $0.1$ to $0.4$. From our experiments, it was clear that varying this number has little effect on the quality of descriptors. A very high number like  $S = 10,000$ still gave good classification results. 
\section{DART-driven Object Classification}
\label{sec:4}
\begin{figure}[t]%
\centering
\includegraphics[width=3in]{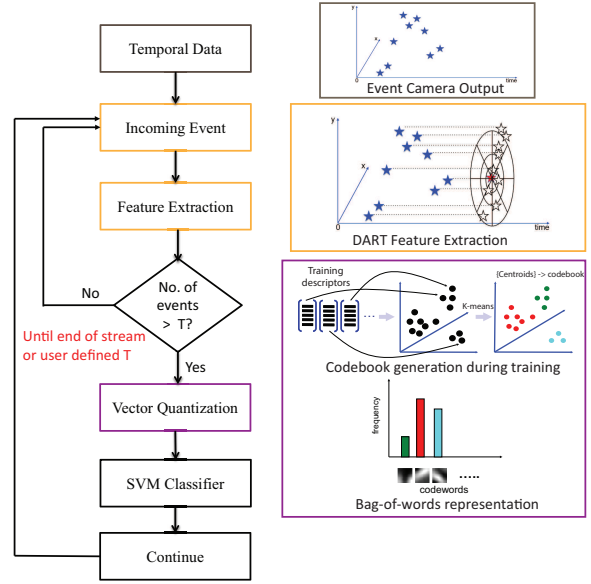}%
\caption{Flowchart of the bag-of-features classification framework (best viewed on a monitor).}%
\label{fig:classify_sc}%
\end{figure}
Temporal difference events from the neuromorphic vision sensors are classified using the standard bag-of-words framework \cite{Csurka04visualcategorization} consisting of four main stages: keypoint selection, feature extraction, vector quantization, and classification. In this work, all events are treated as keypoints and no explicit selection is required. Feature extraction is the process of computing the spatio-temporal descriptors with a log-polar grid with fixed scale and orientation. For the training set, the extracted descriptors are collectively used for K-means to obtain a codebook. The quantization step is the histogram representation of each sample, using the codebook generated in the previous step. Then, the histograms of the training data are used to train a non-linear support vector machine (SVM) classifier. During testing, the codebook construction step is bypassed, and a test sample is simply represented using the codebook and classified using SVM. The flowchart of the classification system is shown in Fig. \ref{fig:classify_sc}.
\par
Each event or feature vector ${\bf{x}}_l$ is quantized into one of $K$ different visual words that are obtained from the training phase. The mapping to a visual word $v_k \in S$ is achieved using a quantization function $f_k({\bf{x}})~:~ S \mapsto [0,1]$. Each quantization function $f_k({\bf{x}})$ is essentially computing the distance of the feature vector to $v_k$ and allowing the assignment if it is minimal. 
\begin{equation}
f_k({\bf{x}}) = f({\bf{x}}; v_k) = I(||{\bf{x}} - v_k|| = \rho)
\label{eq:vectquant}
\end{equation}
where indicator function $I(z)$ outputs 1 when $z$ is true or 0 otherwise; $\rho$ is the Euclidean distance, ${\arg\min_{k}} ||{\bf{x}} - {v_k}||$ . Given K visual words, or K quantization functions $\{f_k({\bf{x}})\}_{k=1}^K$, a codeword representation is computed as,
\begin{equation}
h_j^k = \frac{1}{{{S}}}\sum\limits_{l = 1}^{{S}} {{f_k}({\bf{x}}_l)} 
\label{eq:bow}
\end{equation}
The object representation at a time instance j is expressed by the vector,
\begin{equation}
 {\bf{h}}_j=(h_j^1,h_j^2,\cdots,h_j^K)
\label{eq:BOW}
\end{equation}
\par
Randomized kd-trees are utilized for vector quantization to improve the efficiency of the process in the high-dimensional feature space. Each kd-tree is constructed independently and instead of always splitting on the maximally variant dimension, a random candidate is chosen among the top five most variant dimensions at each level. When querying for the best codeword match, a best-bin-first search is performed across all the trees in parallel.
\par
Nevertheless, the bag-of-words histogram representation discards the spatial relationship between the local features. Therefore, spatial pyramid matching (SPM) \cite{lazebnik2006beyond} is adopted to encode coarse, mid-level spatial relationships between the local features. Subsequently, the SPM object representation is fed to a $\chi^2$-kernel SVM for classification. To efficiently implement the $\chi^2$ kernel, a homogeneous kernel map \cite{Vedaldi2010} is used as a linear approximation. The homogeneous kernel map of order m is a vector function $\Psi (x) \in {\mathbb{R}^{2m + 1}}$ such that the following approximation holds:
\begin{equation}
{k_{{\chi ^2}}}(x,y) \approx \langle \Psi (x),\Psi (y)\rangle
\label{eq:kernelapprox}
\end{equation}
Given the feature map for the scalar case, the corresponding feature map $\Psi (x)$ for the vectorial case is obtained by stacking $\Psi(\mathbf{x}) = ([\Psi ({x_1}), \ldots ,\Psi ({x_m})])$. Note that the stacked feature $\Psi(\mathbf{x})$ has dimension $\left( {d\left( {2m + 1} \right)} \right)$.

\section{Long-term Object Tracking}
\label{sec:5}
The proposed long-term object tracking system consists of two modules: a tracker and a detector. A tracker is a local update for the object while the detector is a global search without spatial constraints. Usually, the tracker gives fast, smooth trajectories of the object, but cannot recover from failure (object lost or occluded) on its own. In which case, an object detector is used to re-initialize the tracker. In the next sections, we present details about the tracker, followed by the detection method.
\subsection{Event-based Object Tracking}
\label{subsec:simpletracker}
In computer vision, object tracking is usually accomplished using a filter-based approach and a simple recognition system is highly unlikely to have a good tracking performance (unless a high very frame rate camera is used). On the other hand, taking advantage of the very high temporal resolution of the event camera, it is possible to extend a recognition system to perform object tracking.
\subsubsection{Training Phase}
\label{sec:tracktrain}
Instead of classifying several object classes against each other, tracking from a classification point-of-view pits the user defined object against every event outside the region-of-interest (ROI), thereby creating a binary classification problem. The task is to ascertain the position of the object, which is contained within the ROI and in turn update the ROI position as long as the object remains in the field-of-view.
\par
To initialize the tracker, the user defines a short time-period and a spatial boundary for the object. Using the DART descriptors within the ROI and outside the ROI, the codebook $C = [v_1, v_2, \cdots, v_K] \in \mathbb{R}^{d \times K} $ is generated in an unsupervised manner as explained in Section \ref{sec:4}. Then the events within the ROI, represented by the set of descriptors $X^{{\omega _1}} = ({\bf{x}}_1, {\bf{x}}_2, \cdots, {\bf{x}}_{C_1})$ can be used to generate a tracker representation ${\bf{h}}_{\omega _1}$ \eqref{eq:BOW}. Similarly, the events outside the ROI $X^{{\omega _2}} = ({\bf{x}}_1, {\bf{x}}_2, \cdots, {\bf{x}}_{C_2})$ can be used for obtaining ${\bf{h}}_{\omega _2}$. However, there are two main concerns: (1) when the ROI descriptors are quantized using the codebook, we end up with a single histogram representation and this is insufficient for training an SVM. Similarly, there is only one data-point after the non-ROI descriptors are quantized; (2) the appearance model generated using a small time-period is insufficient to handle drastic scale and rotation variations of the object. 
\par
To solve the low sample problem, statistical bootstrapping \cite{Wang2005} can be used to generate new subsets of descriptors $\{ X^{{\omega _1}}_1, X^{{\omega _1}}_2, \cdots, X^{{\omega _1}}_{n_1}\}$ and $\{ X^{{\omega _2}}_1, X^{{\omega _2}}_2, \cdots, X^{{\omega _2}}_{n_2}\}$. Specifically, bootstraping $X^{{\omega _1}}$ is the process of random sampling of a subset out of the $C_1$ descriptors, one at a time such that all descriptors have an equal probability of being selected, i.e., $1/{C_1}$. 
\par
Applying bootstrap resampling to the tracking problem, a small number of ROI descriptors are drawn with replacement and quantized to form an SVM data-point. This process is repeated until sufficient number of data-points are generated. Thus, the collection of the bootstrapped representations $\{ {\bf{h}}_{1{\omega _1}},{\bf{h}}_{2{\omega _1}}, \cdots , {\bf{h}}_{{N_1}{\omega _1}} \}$ and $\{ {\bf{h}}_{1{\omega _2}},{\bf{h}}_{2{\omega _2}}, \cdots , {\bf{h}}_{{N_2}{\omega _2}} \}$ can be used to train an SVM classifier. However, an SVM classifier generated using the one-shot learning procedure described above using statistical bootstrapping cannot realistically handle drastic scale and rotation variations of the object. 
\par
In addition to the online SVM update during the tracking phase that ensures the classifier learns such transformations, the key insight is that the DART features extracted after rotation changes of an object are equivalent to the corresponding vertical shifted DART descriptors extracted before the appearance change. In other words, as long as the object boundaries are spatially contained within the maximum radius of the log-polar grid, scale and rotation changes correspond to horizontal and vertical shifts in the log-polar domain \cite{Weiman1979197}. Thus, by inducing random vertical circular shifts to the set of object descriptors $X^{{\omega _1}}$, before quantization, results in bootstrapped representations that are robust to rotation variations.
\par
First, each of the descriptors $({\bf{x}}_1, {\bf{x}}_2, \cdots, {\bf{x}}_{C_1})$ are reshaped to have the original $n_r \times n_w$ DART representation. Then, a random horizontal circular shift factor $f = \lfloor P \times n_w \rfloor$, where $P \sim U([0,1])$, is individually generated and applied to each descriptor. This process is repeated for the non-ROI descriptors to obtain $X_{new}^{{\omega _2}}$. Subsequently, the new set of vectorized descriptors $X_{new}^{{\omega _1}}$ and $X_{new}^{{\omega _2}}$ are bootstrapped for quantization and SVM training.  
\par
A similar approach to tackling scale variations does not work due to its unbounded nature, i.e., rotation changes can be modeled within $[0,2\pi)$ whereas scale change can be fractional or greater than the object size without an upper limit. Nonetheless, the log-polar grid naturally tackles moderate scale variations due to the exponential distribution of the bins. In particular, the bins at the edges of the log-polar grid have a greater spatial resolution, which results in the same distribution as long as the scale change is enclosed by the bin. Therefore, in comparison to a polar or rectangular sampling, log-polar based distribution offers advantageous properties to tackle scale and rotation changes for event-based cameras. 

\subsubsection{Tracking Phase}
\label{sec:tracktest}
\begin{algorithm}[t]
\caption{Event-based Object Tracking}
\label{alg:trackflow}
\textbf{Input}: Initial object state $B_{0}$, codebook $C \in \mathbb{R}^{d \times K}$, tracking rate $e_r$, tracker padding $[p_x,p_y]$ and tracker threshold $\tau_{h}$ \\
\textbf{Output}: Estimated object tracks $B_{t}$
\begin{algorithmic}[1]
\State{Set $track\_flag = 1$, $t=1$, $count=0$ and $fail =0$}
\While{$track\_flag = 1$}
\If{$count = 0$}
\State{Boundary pad $B_{t-1}$ in $(x,y)$ directions by $[p_x,p_y]$ }
\EndIf
\State{Read incoming event $e_i =(x_i,y_i,t_i,p_i)^T$}
\If{$(x_i,y_i) \in B_{t-1}$}
\State{Extract DART descriptor ${\bf{x}}_i$ \eqref{eq:dart_norm} }
\State{Get the quantized representation ${\bf{h}}_t(k)$ as in \eqref{eq:BOW}}
\State{Include $(x_i,y_i)$ in $B_t$}
\State{$count = count + 1$}
\EndIf
\If{($count > e_r \times size(B_{t-1}) $)}
\State{Normalize each entry ${\bf{h}}_t(k) ~ \text{by}~ count$}
\State{Obtain $\Psi({\bf{h}}_t) \in {\mathbb{R}^{2K + 1}} $  \eqref{eq:kernelapprox}} 
\State{Get object score $B^s_{t}$ by SVM projection of $\Psi({\bf{h}}_t)$}
\If{$B^s_{t} < (B_1^s +  \cdots  + B_{t - 1}^s)/(t - 1)$ and $t > 1$}
\State{Failback reset $B_t = B_{t-1}$}
\State{$fail =fail + 1 $}
\If{$fail > \tau_h$} 
\State{Set $track\_flag = 0$ to end tracking}
\EndIf
\Else
\State{Reset $count=0$ and $fail =0$}
\State{Online SVM update using $\Psi({\bf{h}}_t)$ }
\EndIf
\State{Remove empty boundary padding in $B_t$}
\State{Increment object state index $t= t + 1$}
\EndIf
\EndWhile
\State{Output the object tracks $B_{0}, B_{1}, \cdots, B_{t}$}
\end{algorithmic}
\end{algorithm}
Alg.~\ref{alg:trackflow} outlines the proposed event-based tracking method. The initial object state $B_0$, containing the $(x,y)$ locations of the user-specified object events, is boundary padded in $(x,y)$ directions by a few pixels $[p_x,p_y]$. Intuitively, padding ensures that the object motion is captured by the extended region and since event cameras have a high temporal resolution in the order of microseconds, apparent object motion is always smooth in the image plane irrespective of the motion profile. Moreover, the object tracking rate $e_r \in [0,1]$ is specified as a proportion of the object size (simply set to $0.05$ in all our experiments). In other words, the time-period between two track instances is not explicitly chosen and thus tracking is faster when camera motion is faster and vice versa. For the above reasons, padding $[p_x,p_y]$ can be set to the minimum value of one and experiments varying it up to five pixels made no difference to the tracking result. 
\par
Each incoming event within the initial/previous object state $B_{t-1}$ is then used for feature extraction and quantized using the codebook dictionary $C$ to get the object representation ${\bf{h}}_t$ \eqref{eq:BOW}. The normalized object representation ${\bf{h}}_t$ is used to get the higher-dimensional kernel representation $\Psi({\bf{h}}_t) \in {\mathbb{R}^{2K + 1}} $ \eqref{eq:kernelapprox}, which results in better tracking performance with slightly higher computational time compared to a linear SVM as experiments showed. The object score is obtained by projection onto the SVM hyperplane $B_t^s = w^T \Psi({\bf{h}}_t) + b$, where $w$ is the $2K + 1$-dimensional hyperplane and the scalar bias $b$. 
\par
A positive score indicates the presence of the object and a higher value indicates higher confidence of the presence of the object. To ascertain track success, a running average of the SVM scores is used. When $B_t^s$ is lower than the average SVM score, it is likely due to appearance changes over time, especially when the object edge aligns with the camera motion direction resulting in no events registered for that edge. Thus, a fail count is used to keep track of successive instances of lower-than-average object score and when it exceeds $\tau_h$, only then the object is deemed to be lost. On the other hand, a single subsequent instance of $B_t^s > (B_1^s +  \cdots  + B_{t - 1}^s)/(t - 1)$ resets the fail count to zero. Therefore, only when $\tau_h$ continuous occurrences of low object scores results in complete track failure. This strategy automatically accounts for appearance changes while allowing enough instances of online SVM updates to learn a robust model. 
\par
If the SVM classifier score ascertains the object, then the bounding box is updated using the max and min coordinates of the descriptors within the ROI. Subsequently, a non-maximal suppression step removes the padding if there are no events registered in those spatial locations. Note that this tracking algorithm works with less clutter around the object as the non-maximal suppression step greedily chooses a smaller bounding box with the same classification performance to avoid an ever-growing object boundary. Thus, the object is expected to be tracked as long as it remains in the field-of-view of the camera, as outlined in Alg.~\ref{alg:trackflow}. The tougher problem of re-detecting the object when it comes back into the field-of-view of the camera is described in the next subsection. 

\subsection{Event-based Object Detection}
\label{subsec:objdet}
Once the tracker has lost the object, detecting the object is the problem of obtaining a candidate ROI and continuing the tracking process. Therefore, detection is a global search compared to the local sliding window search of the tracker. 
\subsubsection{Training Phase}
After bootstrapping, $p_1 = N_1/(N_1+N_2)$ and $p_2 = N_2/(N_1+N_2)$ is the probability distribution of the training descriptors belonging to the two categories, $\omega_1$ and $\omega_2$ respectively. The codebook construction using k-means partitions the data into K clusters. Then, ${p_{i1}}$ is the ratio of number of samples of class $\omega_1$ in cluster i ($n_{i1}$) to the total number of samples in cluster i ($n_{i}$), i.e., ${p_{i1}} = n_{i1}/n_{i}$. The training phase of the detector is as simple as marking clusters with $p_{i1} > \tau$, and typically set close to 1, say $\tau = 0.95$. Let $d$ denote the number of detector clusters in which the object samples form the majority, and the corresponding cluster indices be $\{k_1,k_2, \cdots, k_d\} ~\text{where}~ d \ll K$. 
\subsubsection{Detection Phase}
Alg.~\ref{ag1} outlines the proposed event-based object detection approach. A detection matrix $M \in \mathbb{R}_+^{h \times w}$ is used to keep track of events that may belong to the object. A binary detection matrix $M_b \in \mathbb{R}_+^{h \times w}$ keeps track of detected events in $M$ with more than $\tau_c$ occurrences. The tracker state $B_t$ is the last tracked ROI location. 
\par
\begin{algorithm}[t]
\caption{Event-based Object Detection}
\label{ag1}
\textbf{Input}: Detector codewords $D$, codebook $C$, object state $B_t$, detector threshold $\tau_d$, image sizes $(h, w)$ and dilation mask $B$ \\
\textbf{Output}: Estimated object state $B_d$
\begin{algorithmic}[1]
\State{Set $detect\_flag = 1$, $B_d = 1_{h,w}$, $\tau_c = 1$ and $count =0$}
\State{Initialize detection matrices $M$ and $M_b$ to $0_{h,w}$}
\While{$detect\_flag = 1$}
\State{Read incoming event $e_i =(x_i,y_i,t_i,p_i,{\bf{x}}_i^T)^T$}
\For{$k=1:K$}
\State{Get the quantization result $f_k({\bf{x}}_i)$ using \eqref{eq:vectquant}}
\If{$f_k({\bf{x}}_i) = 1$ and $k \in D $}
\State{$M(y_i, x_i) = M(y_i, x_i) + 1$}
\State{$count = count + 1$}
\If{$M(y_i, x_i) > \tau_c$}
\State{$M_{b}(y_i, x_i) = 1$}
\EndIf
\EndIf
\EndFor
\If{($count > \tau_d \times h \times w $)}
\State{Perform dilation $M_b \oplus B= \{ z|{(\hat B)_z} \cap {M_b} \ne \emptyset \}$ }
\State{Assign largest connected component in $M_b$ as $B_d$}
\If{$size(B_d) < size(B_t)$ }
\State{Set $detect\_flag = 0$}
\Else
\State{Increase confidence threshold $\tau_c = \tau_c + 1$}
\EndIf
\State{Reset $M = 0_{h,w}$, $M_b = 0_{h,w}$, and $count =0$ }
\EndIf
\EndWhile
\State{Output the coordinates in $B_d$ as the object state}
\end{algorithmic}
\end{algorithm}
For every incoming event, the quantization function \eqref{eq:vectquant} determines whether it belongs to the detector clusters $\{k_1,k_2, \cdots, k_d\}$ and updates $M$ and $M_b$. The detector threshold $\tau_d \times h \times w ~\text{where} \tau_d \in [0,1]$ determines if enough events have been accumulated within the detection matrix $M$. The parameter $\tau_d$ is set to 0.25, meaning at least 25\% of the events have occurred for the detection process.  
\par
A binary dilation is then performed on $M_b$ to connected neighboring pixels if they have a detected event. For pixel $M_b(y,x)$, the four neighboring pixels are $M_b(y,x-1)$, $M_b(y-1,x)$, $M_b(y,x+1)$, and $M_b(y+1,x)$. Two object pixels $p$ and $q$ are said to be \emph{4-connected} if there is a path which consists of object pixels $a_1, a_2, \cdots, a_r$ such that $a_1 = p$ and $a_2 = q$, and for all $1 \leq m \leq r-1$, $a_m$ and $a_{m+1}$ are 4-neighbor for each other. The dilation pixel ensures that object pixels are not isolated. 
\par
The largest 4-connected component in $M_b$ is a candidate ROI, denoted by $B_d$. If the area of $B_d$ is larger than the last tracker state $B_t$, then detection is performed again with an incremented confidence threshold $\tau_c$. In other words, object events have a higher probability of appearing more number of times in $M$, and thus the detected area will reduce in size after thresholding, $M(y_i, x_i) > \tau_c$. During this process, the parameter $\tau_c$ is set automatically based on the size of the detected ROI. 
\subsection{Event-based Tracking and Detection}
\label{sec:lot}
The event-based long-term object tracking (eLOT) framework combines the tracker (Sec.~\ref{subsec:simpletracker}) and the detector (Sec.~\ref{subsec:objdet}) to track a desired object indefinitely. Since the global search of the detector is computationally intensive, it is activated only when the local search tracker fails. During the tracking process, online learning is needed to account for the changes in object appearance as discussed earlier. In particular, the binary classifier used by the tracker is updated when the region-of-interest (ROI) is classified as the object, and updating the tracker mitigates the drifting issue. It is worth noting that the tracker is updated only when the tracking confidence $B_t^s$ is higher than the mean tracking score. Also note that online update for the detector is not present in the current setup, because of the computationally intensive \emph{k}-means needed for updating the codebook and re-learning the complete setup consisting of the tracker classifier and detector. 
\par
Being one of the first works in this research domain, the objective of this work is to set a baseline for further extensive research in event-based tracking, and thus we did not fine-tune parameters extensively for achieving better performance. In fact, the lower limits of intuitive parameters like the tracking event rate $e_r$ and detector threshold $\tau_d$ can be easily set as a proportion of object or image sizes. Furthermore, other parameters like the tracker object score $B_t^s$ and the detector confidence $\tau_c$ are set on-the-fly automatically and need no fine-tuning. 

\section{DART-based feature matching}
\label{sec:6}
Feature matching is a fundamental aspect of many problems in vision, including object or scene recognition, stereo correspondence, and motion tracking. Recognizing previously encountered scenes is an important addition to extent the ability to obtain visual odometry from pure event streams \cite{Rebecq2017}. In other words, when the feature descriptors are distinctive, a set of new features can be correctly matched with high probability against a large database of previously seen features, providing a basis for a full-scale SLAM system with loop closure.
\par
Traditionally, feature matching in computer vision is performed between two sets of descriptors extracted from two different frames. For silicon retinas, feature matching is still done between two sets of descriptors, but obtained from different spatio-temporal slices. For large feature sets, it is imperative to use an efficient approximate nearest neighbor search. Like \cite{lowe1999object}, we use a distance ratio threshold, in the range $(0,1]$, for rejecting ambiguous matches. If the ratio of the distances to the first nearest neighbour and second nearest is less than 0.6, a feature descriptor is said to have a good match with the nearest neighbour in the second set.

\begin{figure*}
\centering
\subfloat[Anchor]{\includegraphics[width=0.15\textwidth]{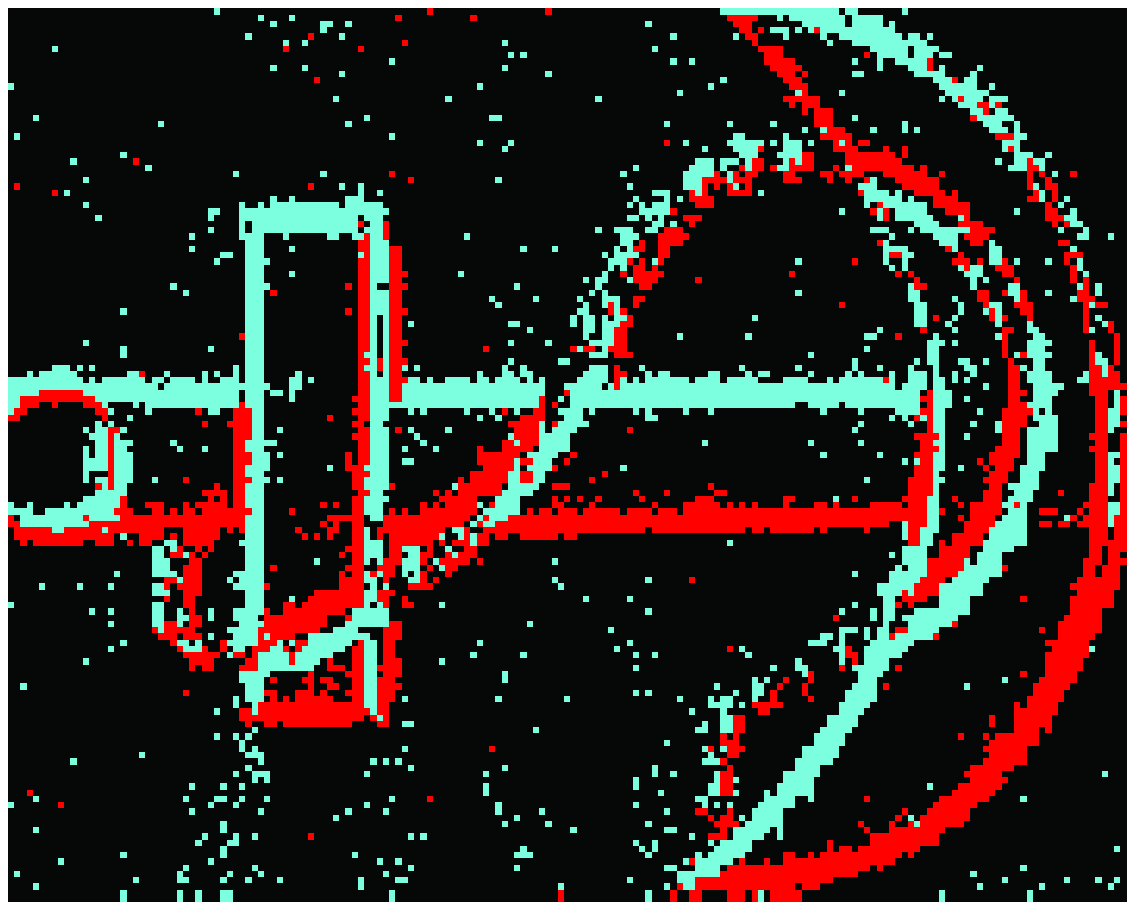}%
\label{fig_first_case}}
\hspace{1em}%
\subfloat[Binocular]{\includegraphics[width=0.15\textwidth]{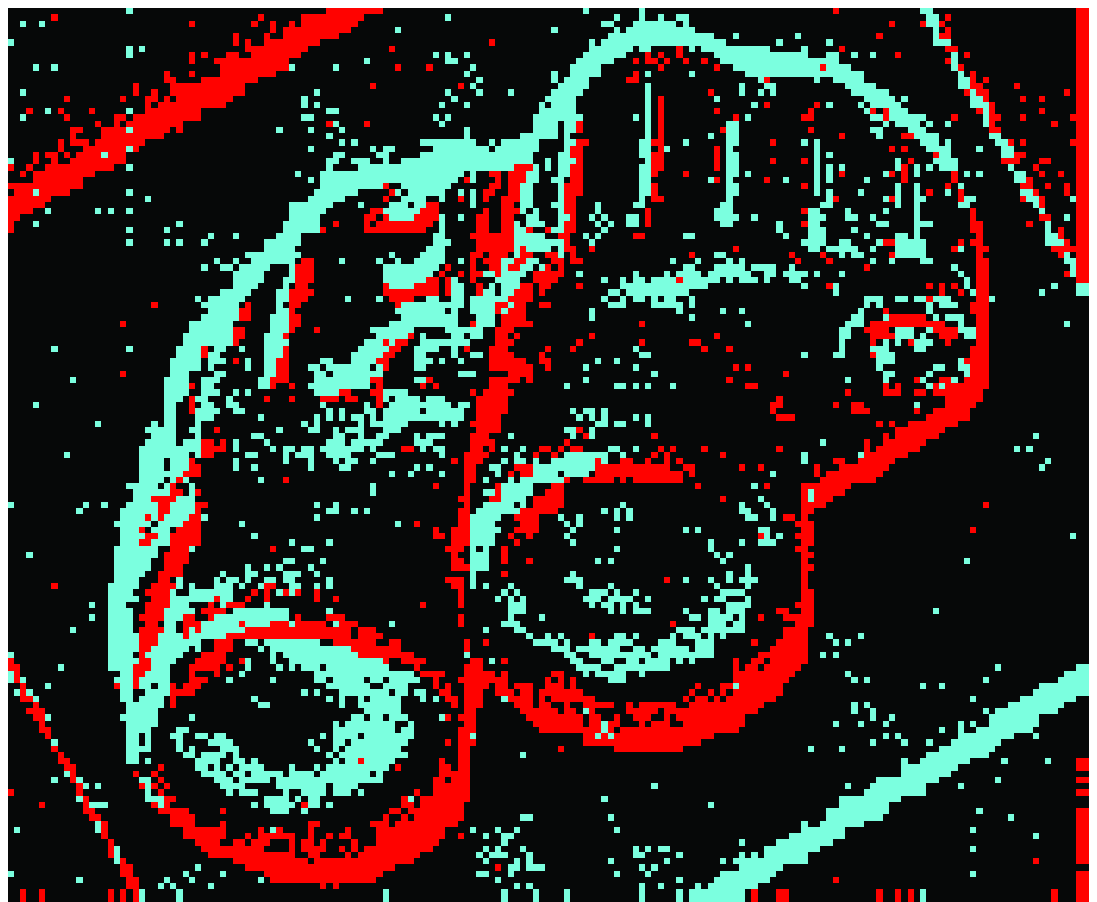}%
\label{fig_second_case}}
\hspace{1em}%
\subfloat[Buddha]{\includegraphics[width=0.15\textwidth]{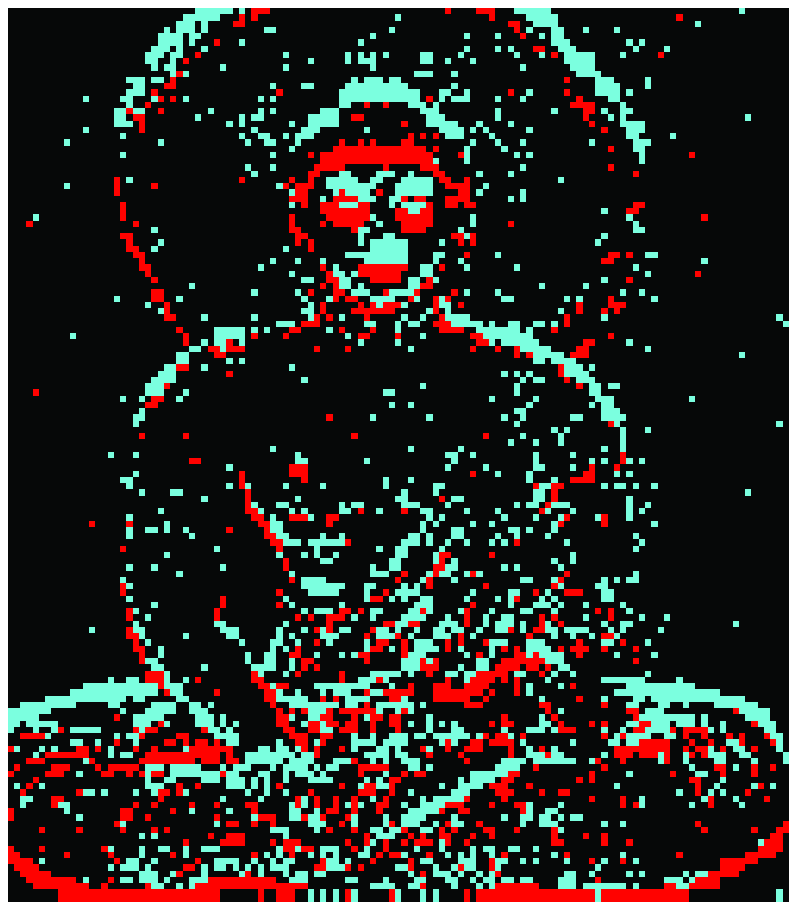}%
\label{fig_third_case}}
\hspace{1em}%
\subfloat[Crayfish]{\includegraphics[width=0.15\textwidth]{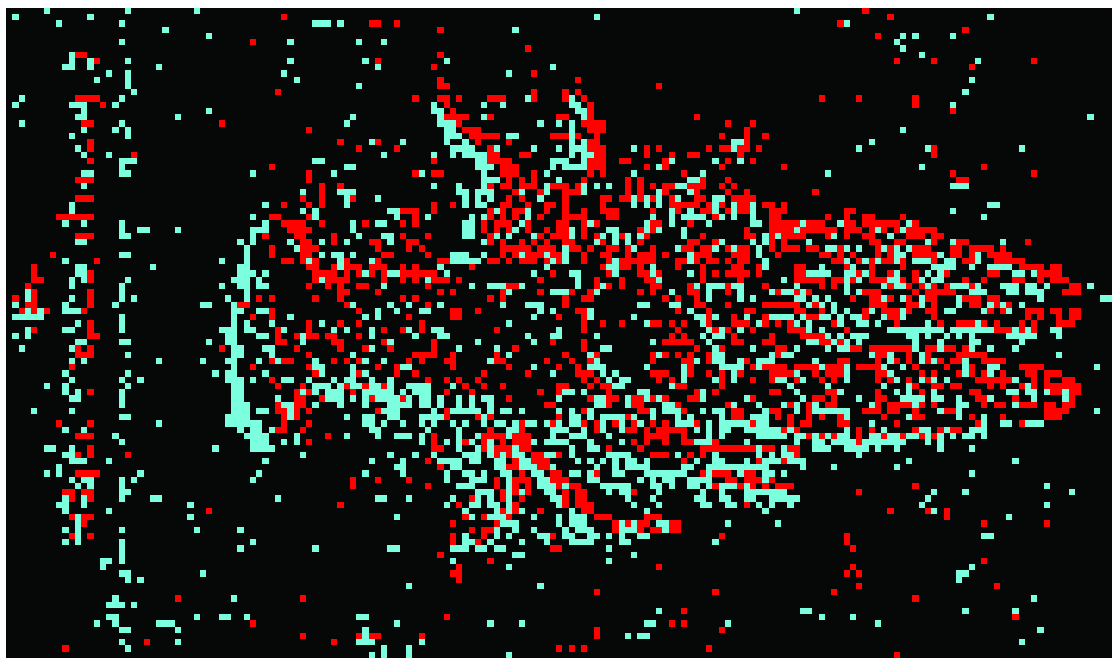}%
\label{fig_ssecond_case}}
\hspace{1em}%
\subfloat[Dragonfly]{\includegraphics[width=0.15\textwidth]{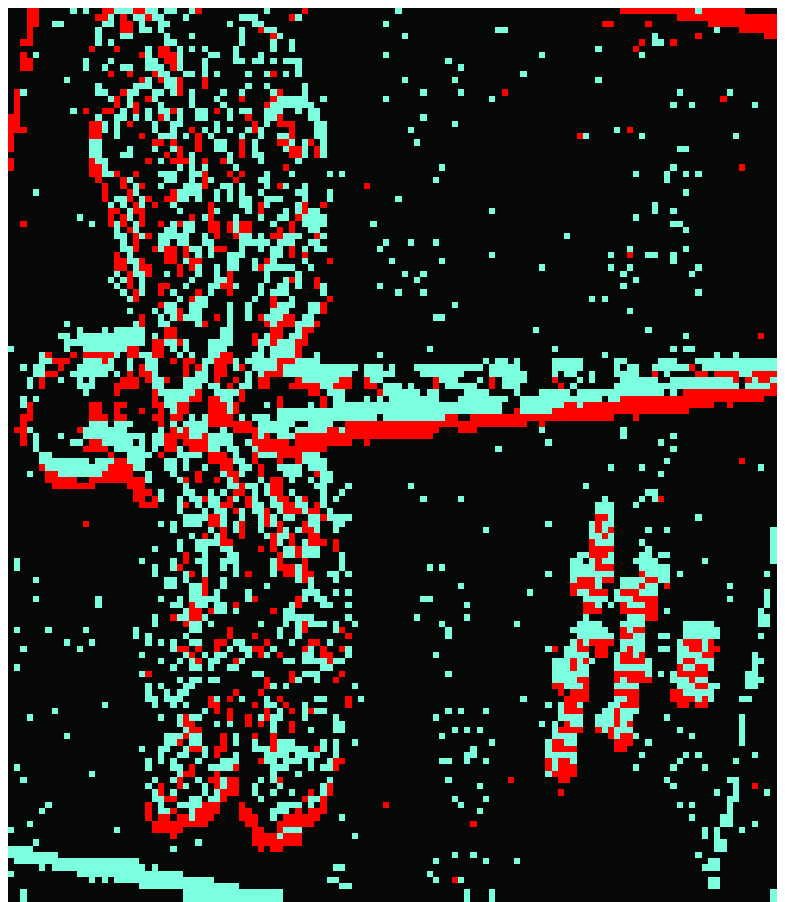}%
\label{fig_fourth_case}}
\hspace{1em}%
\subfloat[Gramophone]{\includegraphics[width=0.15\textwidth]{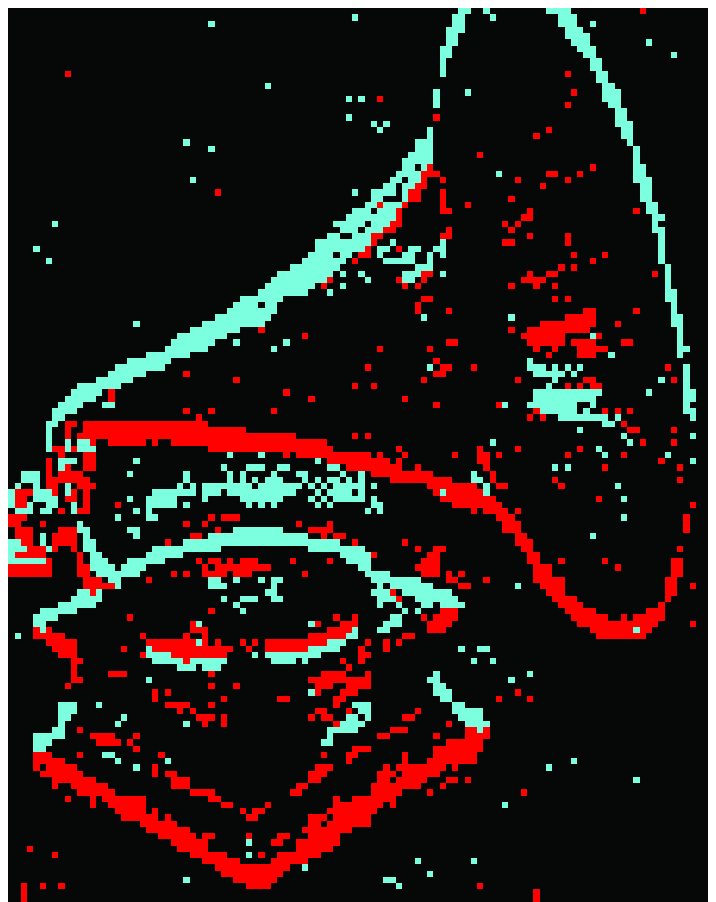}%
\label{fig_fifth_case}}
\hspace{1em}%
\subfloat[Laptop]{\includegraphics[width=0.15\textwidth]{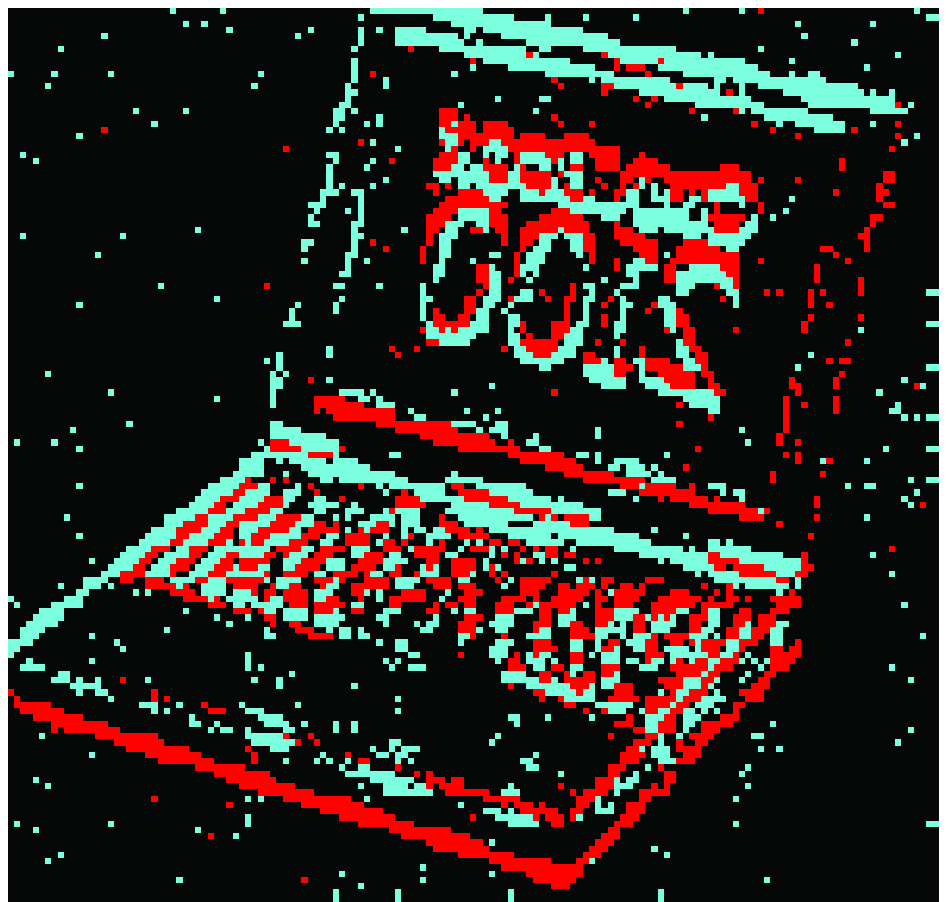}%
\label{fig_sixth_case}}
\hspace{1em}%
\subfloat[Pyramid]{\includegraphics[width=0.15\textwidth]{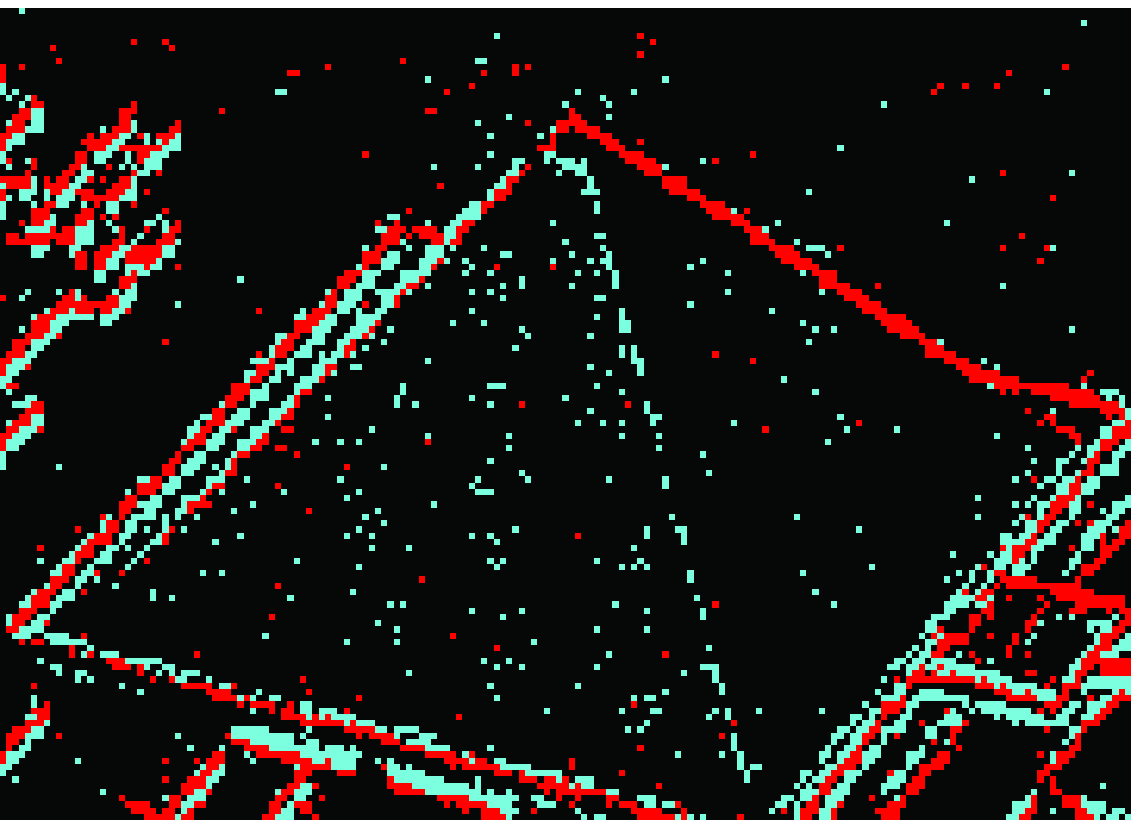}%
\label{fig_seventh_case}}
\hspace{1em}%
\subfloat[Snoopy]{\includegraphics[width=0.15\textwidth]{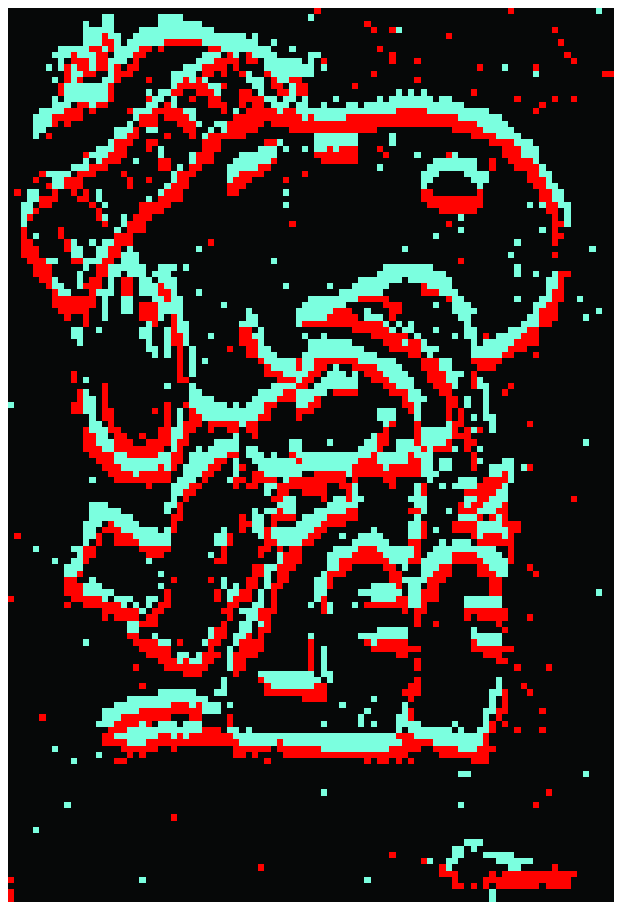}%
\label{fig_eighth_case}}
\hspace{1em}%
\subfloat[Stop Sign]{\includegraphics[width=0.15\textwidth]{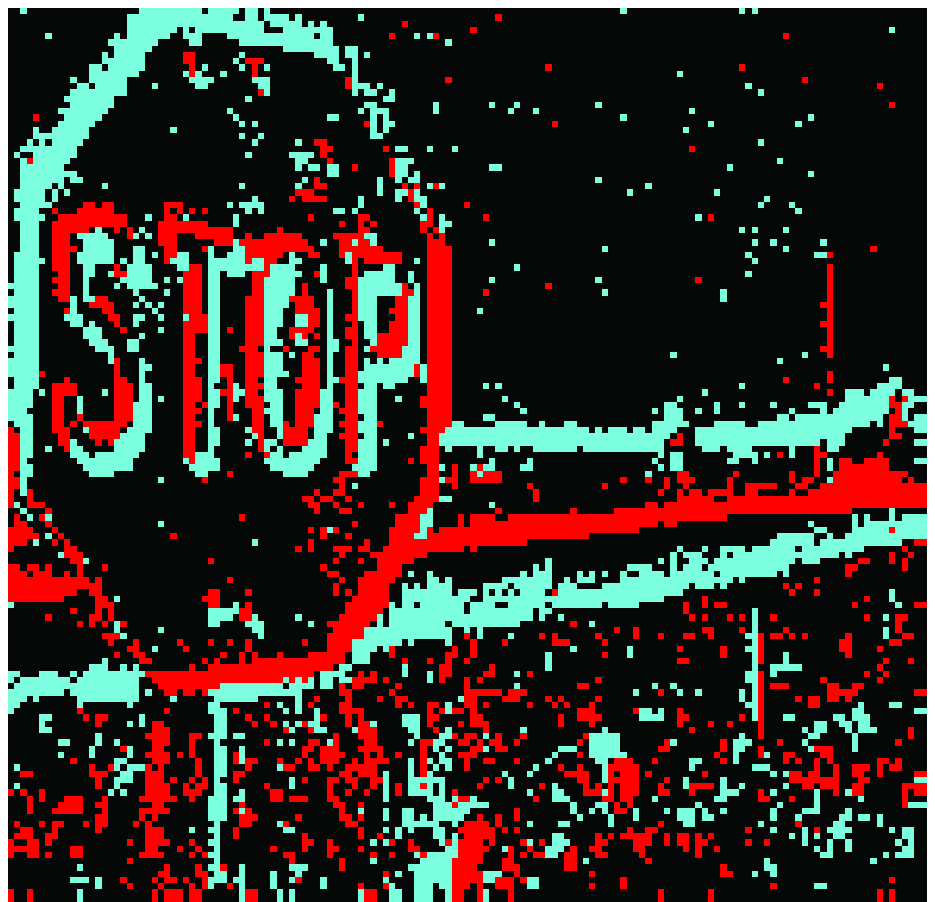}%
\label{fig_ninth_case}}
\hspace{1em}%
\subfloat[Watch]{\includegraphics[width=0.15\textwidth]{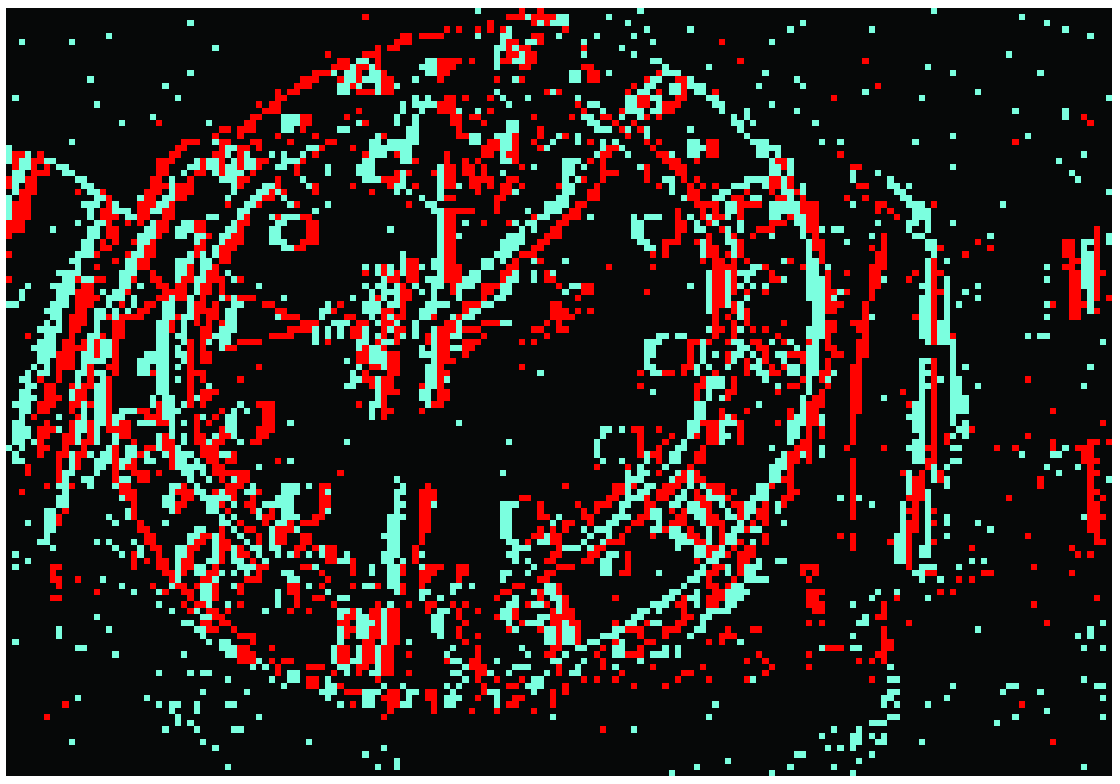}%
\label{fig_tenth_case}}
\hspace{1em}%
\subfloat[Yin Yang]{\includegraphics[width=0.15\textwidth]{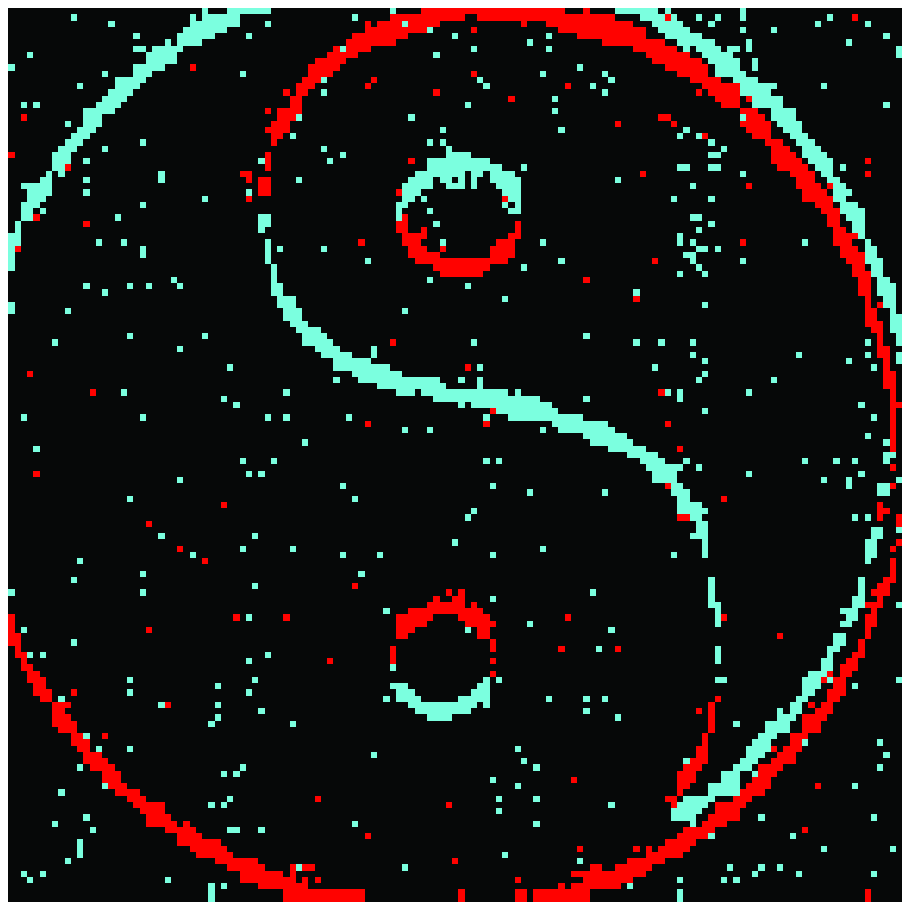}%
\label{fig_tenthd_case}}
\hspace{1em}%
\caption{Samples from the N-Caltech101 dataset where red and cyan colorization represent the polarity (brightness increase/decrease) of the event.}
\label{fig:ncaltech}
\end{figure*}
\section{Experimental Setup}
\label{sec:7}
\subsection{Object Classification Datasets}
We tested the proposed object classification framework (Sec.~\ref{sec:4}) on five neuromorphic vision datasets.
\subsubsection{N-MNIST}
We first tested the proposed classification system on the N-MNIST database introduced by Orchard~et~al.~\cite{Orchard2015}, which consists of $60,000$ training samples and $10,000$ testing samples from 10 categories. The object classes are digits 0-9 and the dataset is based on converting the original MNIST dataset using a pan-tilt camera unit and an image projector. The dataset recordings are stabilized before classification.
\subsubsection{MNIST-DVS}
Next, we tested our framework on the MNIST-DVS \cite{Serrano-Gotarredona2013} using the protocol followed by previous works \cite{Peng2017, Li2016}. The MNIST-DVS dataset contains three scales of the digits each having $10,000$ samples. We follow previous works by using scale 4 of the dataset, and performing training with 90\% of the samples chosen randomly and testing with the remaining $1,000$ samples. This experiment is repeated 10 times and the average accuracy is reported.
\subsubsection{N-Caltech101}
The N-Caltech101 dataset \cite{Orchard2015}, which is a spiking version of the original frame-based Caltech-101 dataset, is to date one of the most challenging the neuromorphic vision datasets (Fig.~\ref{fig:ncaltech}). It consists of 101 object categories\footnote{The N-Caltech101 dataset does not contain the ``Faces" class in order to avoid confusion with the ``Faces\_easy" class, leaving 100 object classes plus a background class.} with varied number of recordings in each category (ranging from 31 to 800 samples). We follow the standard experimental protocol for this dataset \cite{lazebnik2006beyond, Ramesh2017}, which is to train on 30 images and test with a maximum of 50 images per category.
\subsubsection{CIFAR10-DVS}
We report the classification results on the recently introduced neuromorphic vision dataset, CIFAR10-DVS, which contains a total of 10,000 event-stream recordings in 10 classes (airplane, automobile, bird, cat, deer, dog, frog, horse, ship, truck) with 1000 recordings per class. The testing protocol for this dataset is the same as MNIST-DVS. Note that CIFAR10-DVS dataset recordings were not stabilized before classification.
\subsubsection{N-SOD}
The in-house Neuromorphic Single Object Dataset (N-SOD) is collected for the purpose of testing object recognition under different view-points, and is also used for developing a real-time object recognition system. The dataset contains three object categories with samples of varying length in time (ranging from 5 s to 20 s). The three objects to be recognized are a thumper 6-wheel robot (Dagu Wild Thumper 6WD All-Terrain Chassis with a RoboClaw controller), an unmanned aerial vehicle, a box (assumed to be an obstacle  with some printed signs) along with a background class. 
\subsection{Dataset for tracking}
For testing the proposed eLOT system, the shapes data in the event-camera dataset \cite{Mueggler2017} was used to track object shapes. For each object shape, the training ROI was manually specified during the first 300ms of the recording and the testing was done up to end of the recording (60s). Using the ground truth annotations we created, it is possible to quantitatively evaluate the tracking performance and this sets up one of the first tracking benchmark for the neuromorphic vision community. The object location is specified as a bounding box within a short time-interval of 10ms for the full-length of the shapes data. 
\par
In general, tracking algorithms are evaluated by three metrics \cite{wu2013online}, which are center location error (CLE), overlap success (OS) and distance precision (DP). The first metric, CLE, indicates the average Euclidean distance between the ground-truth and the estimated center location. The second metric, OS, is defined as the percentage of times the bounding box overlap with ground truth surpasses a threshold. However, the third metric, DP, is the percentage of frames whose estimated location is within the given threshold distance of the ground truth, which is less applicable for frame-less event cameras. Moreover, for long-term tracking problem, the annotations for certain time-intervals may be empty and thus we only use OS as the primary metric for our evaluation and we report the results at a threshold of 0.5, which correspond to the PASCAL evaluation criteria. In addition, we also report CLE when there is an overlap success to show the closeness of ground truth match.  
\par
However, the OS metric does not directly penalize false positives, i.e., if the tracked result is the whole image plane, then OS metric gives a perfect track score. Thus, we also report the intersection over union score (IoU) \cite{Nowozin2014}, defined as follows,
\begin{equation}
IoU = \frac{TP}{TP + FP + FN}
\label{eq:iou}
\end{equation}
where TP, FP and FN denote true positive, false positive and false negative. In the context of event cameras, the ground truth bounding box location within a short time interval is used to count the number of events that match the ground truth events to calculate the TP, FP and FN. Similarly, OS is calculated as the percentage of events that overlaps with the ground-truth events within a time-interval. 
\section{Results}
\label{sec:8}
\subsection{Object Classification}
\label{subsec:objclass}
For the DART descriptor, a 7 by 12 log-polar grid \cite{Ramesh2017b} with a minimum radius of 2 pixels \cite{shapecontext_belongie} and a maximum radius set to 10 pixels was used. A $1\times1$, $2\times2$, and a $3\times3$ spatial pyramid representation is used to pool the descriptors before classification using the SVM. Normalization is done at each level of SPM, before doing a final normalization for the entire representation. For instance, the four bag-of-words histograms from the $2 \times 2$ grid are normalized separately and concatenated together with the normalized representation from the other levels. All the descriptors in each recording of the dataset are used for classification, but we also report temporal classification performance for the N-MNIST dataset. In other words, each 300 ms N-MNIST recordings is classified every 10 ms and a majority voting result is reported. In this work, we used a codebook size of $3,000$ for all the reported results. The source code for the classification framework can be downloaded from the Bitbucket repository\footnote{Source code: https://goo.gl/cVNWLB (with tracking annotations)}, which also contains the object annotations for the tracking evaluation. 
\subsubsection{N-MNIST}
\label{sec:results_objclass}
\begin{figure}[t]
\centering
\includegraphics[width=3in]{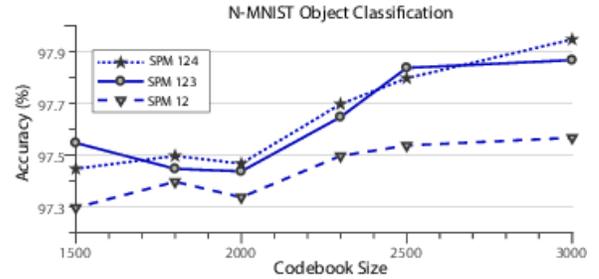}%
\caption{Codebook size and SPM grid Vs. Accuracy.}%
\label{fig:codebookvary}%
\end{figure}

\begin{table}[t]
\begin{center}
\caption{
Classification accuracy on N-MNIST and N-Caltech101 datasets (\%).
}
\label{table:accuracy}
\begin{tabular}{lcc}
\hline\noalign{\smallskip}
~ & {\bf N-MNIST} & {\bf N-Caltech101} \\
\noalign{\smallskip}
\hline
\noalign{\smallskip}
H-First \cite{Orchard2015a}  & 71.20 & 5.40\\
HOTS \cite{Lagorce2016} & 80.80 & 21.0\\
Gabor-SNN \cite{Sironi2018} & 83.70 & 19.60\\
HATS \cite{Sironi2018} & {\bf 99.10} & 64.20\\
DART & 97.95 & \textbf{66.42}\\
DART (with object outline) & - & \textbf{70.33} \\
\hline
Phased LSTM & 97.30 & -\\
Deep SNN & 98.70 & -\\
\hline
\end{tabular}
\end{center}
\end{table}
\setlength{\tabcolsep}{1.4pt}
Table~\ref{table:accuracy} shows the accuracy of the proposed method on the N-MNIST object classification challenge compared to recent works. Using the DART descriptors in a bag-of-words framework, as described in Section \ref{sec:4}, we can obtain a competitive result compared to deep neural network schemes. Fig.~\ref{fig:codebookvary} shows the effect of varying the codebook size on classification accuracy for different SPM grid parameters. As expected, higher codebook sizes lead to better classification accuracy \cite{Nowak2006} and the standard three level SPM grid representation performs better than the two level representation. 
\par
\begin{figure}[t]
\centering
\includegraphics[width=2.5in]{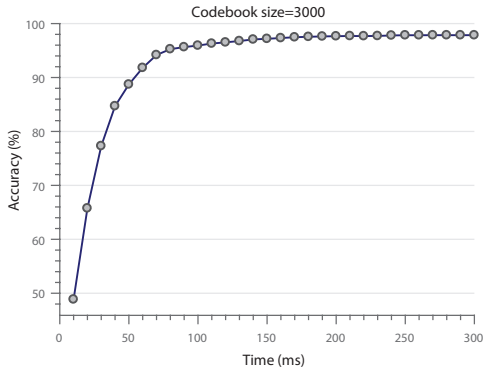}%
\caption{Classification of the NMNIST samples at regular time intervals.}%
\label{fig:timevary}%
\end{figure}

In addition to the results obtained using the entire recording of each N-MNIST sample, the temporal sequence can be classified at regular intervals by updating the pooled bag-of-words representation for a fixed length of time and performing classification with the intermediate histogram. Fig.~\ref{fig:timevary} shows the result of classifying at regular intervals of each N-MNIST sample of the test set. It is evident that as more information flows into the bag-of-words representation, the accuracy increases and when it reaches around 150 ms, the accuracy reaches very close to the best result of using all the 300 ms of information in each sample.
\subsubsection{MNIST-DVS}
\label{sec:results_mnistclass}
\begin{table}[t]
\begin{center}
\caption{
Classification accuracy on MNIST-DVS and CIFAR10-DVS dataset (\%).
}
\label{table:accuracy2}
\begin{tabular}{lcc}
\hline\noalign{\smallskip}
~ & {\bf MNIST-DVS} & {\bf CIFAR10-DVS} \\
\noalign{\smallskip}
\hline
\noalign{\smallskip}
H-First \cite{Orchard2015a}  & 59.50 & 7.70\\
HOTS \cite{Lagorce2016} & 80.30 & 27.10\\
Gabor-SNN \cite{Sironi2018} & 82.40 & 24.50\\
Peng's \cite{Peng2017} & 76.49 $\pm$ 11.77 & 31.01 \\
HATS \cite{Sironi2018} &  98.40 & 52.40\\
DART & \textbf{98.51 $\pm$ 0.30} & \textbf{65.78}\\
\hline
Zhao's \cite{Zhao2015} & 75.52 $\pm$ 11.17 & - \\
Random Forest \cite{Li2016} & 88.39 $\pm$ 1.54 & -  \\
\hline
\end{tabular}
\end{center}
\end{table}
\setlength{\tabcolsep}{1.4pt}
\begin{figure}[t]
\centering
\includegraphics[width=3in]{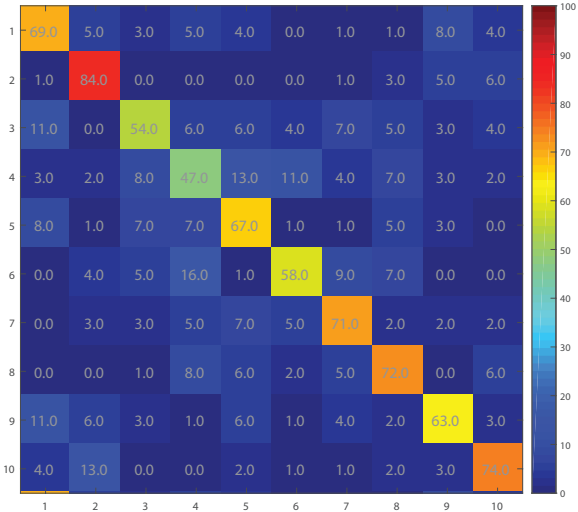}%
\caption{Confusion matrix for the best result on the CIFAR10-DVS dataset in which the rows denote the actual class and columns represent the predicted class.}%
\label{tab:CIFARDVSconfmat}%
\end{figure}
\begin{figure}%
\centering
\includegraphics[width=2.5in]{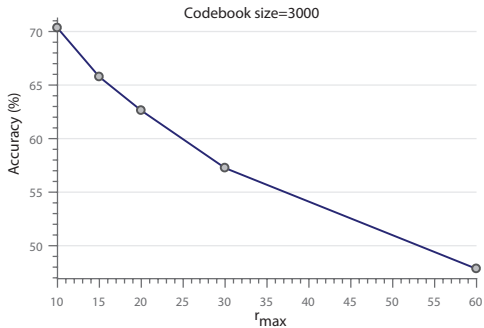}%
\caption{Effect of varying the maximum radius of the log-polar grid on NCaltech-101 classification accuracy}%
\label{fig:rmax_acc}%
\end{figure}
Table~\ref{table:accuracy2} shows the performance of DART in comparison to state-of-the-art methods on the MNIST-DVS dataset. On the challenging MNIST-DVS, whose recordings have noise, blur and other factors caused by a fixed AER DVS used to capture the moving digit images, the superiority of the DART-based classification is clear compared to existing formulations. Note that the same set of parameters used in reporting the results on N-MNIST was also applied to the MNIST-DVS dataset. 
\par
In order to further test the scale robustness of the proposed DART descriptor, we use the different scale recordings in the MNIST-DVS dataset. Specifically, we used $10,000$ samples each from scale-4 and scale-8 for training and $10,000$ samples of scale-16 for testing. The proposed DART-driven classification achieved an accuracy of $24.42\%$, which is reasonable considering the log-polar grid parameters used were the same for all scales. In fact, it is well-known in the computer vision community that gradual scale variations can be easily handled using log-polar grids \cite{shapecontext_belongie, Ramesh2017b}, whereas the MNIST-DVS dataset contains samples that are very different in scales. This is a plausible explanation for the reported classification accuracy, which in any case is the first and the highest result using the different scales of the MNIST-DVS dataset. 
\subsubsection{CIFAR10-DVS}
CIFAR10-DVS is a relatively new dataset compared to N-MNIST or N-Caltech101. In comparison to existing works as shown in Table.~\ref{table:accuracy2}, the proposed method obtains $65.43 \pm 0.35 $ on the ten classes of the CIFAR10-DVS dataset. Note that the same set of parameters used in reporting the results on the previous datasets was also applied. Fig.~\ref{tab:CIFARDVSconfmat} shows the confusion matrix for the best result of the ten trials used in the experimental protocol.
\subsubsection{N-Caltech101}
The best result reported for N-Caltech101 in the literature is 64.20\% using the unpublished Histogram of Averaged Time Surfaces (HATS) approach \cite{Sironi2018}. In contrast, the proposed method in this paper using DART achieves an accuracy of 66.42\%, and a corresponding weighted accuracy of 65.6\% which is calculated by averaging the individual class accuracies. 
\par
The above results highlight the ability of the DART descriptor to capture precise spatio-temporal information. As the confusion matrix containing 101 object categories is impractical to list, we report some of the best and worst performing categories by our classification system. There were 18 categories that had more than 90\% of the test images classified correctly, namely Faces\_easy (47/50), Motorbikes (50/50), accordion (24/25), airplanes (47/50), binocular (2/3), car side (47/50), dalmatian (34/37), dollar bill (22/22), ewer (47/50), Garfield (4/4), inline\_skate (1/1), metronome (2/2),  minaret (45/46), pagoda (16/17), scissors (9/9), snoopy (5/5), trilobite (50/50), and windsor chair (46/46), where the numbers in the brackets indicate the number of wrongly classified images and the total number of test images. It is worth mentioning that categories like faces, airplanes and side view of cars have been noted as easy to classify in the Caltech-101 too.
\par
There were a few classes with more than 60\% of their test images classified wrongly: Background (5/50), anchor (4/12), brontosaurus (5/13), cougar body (5/17), crab (17/43), crocodile (2/20), emu (4/23), flamingo head (5/15), gerenuk (1/4), ibis (12/50), water lily (2/7), and wild cat (1/4). With the exception of the background class and anchor, it is interesting to note that non-rigid objects like animals are difficult to classify, since the intra-class pose and appearance variations are expected to be very high.
\par
While the Neuromorphic-Caltech101 dataset is an exact spiking version replica of the original frame-based Caltech101 dataset, it additionally comes along with annotations for the objects for each recording. Contrary to the role of context/background to object recognition using standard cameras \cite{Galleguillos2010}, the event stream does not carry any color information, which is one of the primary visual cues for separating foreground from background \cite{Livingstone1987,Xie2011}. This implies background information may hinder accurate event-based object recognition, especially in the face of clutter. We verified this by making use of object outlines provided in the N-Caltech101 dataset.
\par
The classification accuracy using the annotations is significantly higher at 70.33\%. Interestingly, the object classes with the worse performance were quite different to the earlier setup without the annotations. Ant, beaver, cannon, cougar body, crab, crocodile, emu, llama, mayfly, octopus, platypus, scorpion, starfish, water lily and wrench were the categories with more than 60\% of their test images classified wrongly. The categories with more than 90\% of the test images classified correctly remained largely unaffected with a couple of additions. 
\par
Since the NCaltech-101 dataset is closer to real-world recordings compared to the other datasets used in this work, we test the performance of the feature parameters. One of the most important parameters of the DART descriptor is the maximum radius of the log-polar grid, which determines the contextual information captured by the descriptor.
\par
Fig.~\ref{fig:rmax_acc} shows the effect of varying the maximum radius of the log-polar grid on the testing classification accuracy. This is a contrasting result, compared to the log-polar grid performance on binary images where performance increases with increase in the radius of the grid \cite{shapecontext_belongie}. However, the trend is similar to the log-polar grid performance on gray-scale images \cite{Kokkinos2008, Ramesh2017}. Thus, we set the maximum radius of the DART grid to be 10 pixels, for the real-time experiments reported below.

\subsubsection{N-SOD}
\label{sec:recog_results_inhouse}
Testing on N-SOD was carried out mainly to create a real-time object classification system in an indoor laboratory setup. Firstly, recordings of individual objects under different appearance variations were collected to form the N-SOD dataset , which was used for offline training \& validation using the DART-driven object classification framework (Sec.~\ref{sec:4}). Subsequently, an event camera on-board a hovering UAV in the indoor setting is expected to identify the objects in its field-of-view using the trained model. Due to the physical limitation of the UAV, we also performed hand-held testing to showcase robust classification.
\par
For the purpose of real-time implementation, the classification framework was implemented in C++ and when combined with a ROS interface in Ubuntu, live data from the DAVIS camera can be processed in real-time. A demo can be viewed here \footnote{https://youtu.be/8SeoJurs-tk}, which showcases a hovering UAV fitted with a downward-looking event camera for recognizing objects on the floor and also to indicate if no objects are present with a background tag. The processing is done on-board using a Compute Stick with an Intel Core m5-6Y57 vPro processor. The purpose of this demo is to showcase real-time processing using the DART-driven object classification method presented in Sec.~\ref{sec:4}. 
\par
The hand-held experiments with a freely moving event camera can be viewed here\footnote{https://youtu.be/jKJF\_g73jAo}. For ease of viewing, the video is edited to show glimpses of the experiments highlighting scale, rotation, view-point and occlusion. The video description contains the full-length video for a closer look. It is clear from these experiments that the proposed DART-driven classification method can handle drastic appearance changes. An important point to note is that the training data contains only some of the variations tested and the training samples are without occlusions under normal lighting conditions. 
\par
The processing rate of the descriptor computation using a single-threaded C++ implementation was about 59000 events/sec on an Intel i7 vPro desktop. Although this number is modest, it is straightforward to boost the performance on neuromorphic processors/FPGA with parallel computing abilities or simply by adopting a multi-threaded implementation. The focus of this manuscript is to show new capabilities using these novel sensors without exclusive focus on real-time performance, which is an important future research direction. 

\subsection{Object Tracking Results}
\label{sec:tracking_det}
\begin{table}[t]
\begin{center}
\caption{Quantitative tracking results using eLOT on the event camera dataset.}
\label{tab:shapetrack}
\begin{tabular}{lccc}
\hline\noalign{\smallskip}
Translation & {\bf OS} & {\bf CLE} & {\bf IoU} \\
\noalign{\smallskip}
\hline
\noalign{\smallskip}
Hexagon   &	0.6414  &	0.0717 &	0.5909 \\
Triangle	& 0.7619	& 0.0602 &	0.7163 \\
Lshape	  & 0.5933	& 0.0833 &	0.5558 \\
Star	    & 0.4976	& 0.0301 &	0.4869 \\
Oval	    & 0.4506	& 0.0685 &	0.4136 \\
Rectangle	& 0.7437	& 0.1295 &	0.6768 \\
Heart	    & 0.6811	& 0.0724 &	0.6428 \\
\hline
\textit{Average} & \textbf{0.6242} & \textbf{0.0737} & \textbf{0.5833} \\ 
\hline
\\
\end{tabular}

\begin{tabular}{lccc}
\\
\hline\noalign{\smallskip}
Rotation & {\bf OS} & {\bf CLE} & {\bf IoU} \\
\noalign{\smallskip}
\hline
\noalign{\smallskip}
Hexagon  & 	0.6427	& 0.0674 &	0.5544 \\
Triangle &	0.8795	& 0.1014 & 	0.7981 \\
Lshape	 & 0.5653	  & 0.1304 &	0.4972 \\
Star	   & 0.7029	  & 0.0996 &	0.6787 \\
Oval	   & 0.5829	  & 0.1040 &	0.4976 \\
Rectangle&	0.5820	& 0.2127 &	0.5317 \\
Heart	   & 0.5546	  & 0.1183 &	0.4903 \\
\hline
\textit{Average} & \textbf{0.6443} & \textbf{0.1191} & \textbf{0.5783} \\  
\hline
\\
\end{tabular}

\begin{tabular}{lccc}
\\
\hline\noalign{\smallskip}
6-DOF & {\bf OS} & {\bf CLE} & {\bf IoU} \\
\noalign{\smallskip}
\hline
\noalign{\smallskip}
Hexagon  & 	0.6608	& 0.0600 &	0.6515 \\
Triangle &	0.7004	& 0.0623 & 	0.6936 \\
Lshape	 & 0.7138	  & 0.0651 &	0.6610 \\
Star	   & 0.2461	  & 0.0228 &	0.2480 \\
Oval	   & 0.6906	  & 0.0884 &	0.6297 \\
Rectangle&	0.6726	& 0.0979 &	0.6254 \\
Heart	   & 0.4401	  & 0.0418 &	0.4025 \\
\hline
\textit{Average} & \textbf{0.5892} & \textbf{0.0626} & \textbf{0.5588} \\ 
\hline
\end{tabular}
\end{center}
\end{table}
\setlength{\tabcolsep}{1.4pt}
Seven shapes from the event-camera dataset \cite{Mueggler2017} are used to test the tracking system, namely triangle, star, hexagon, oval, rectangle, heart and an L-shape with three different camera motion profiles: translation, rotation and 6-DOF. For all the experiments, a small codebook size of 300 was used to model the ROI and the background without spatial pyramid matching. This is done for simplicity and to achieve a faster classification result. In addition, an approximate nearest neighbour scheme is adopted by limiting the maximum number of codeword distance comparisons to 15 for faster vector quantization using the \emph{k}-d tree. The SVM training is performed with a bootstrapping that outputs equal number of samples as the initial number of descriptors. 
\par
It is interesting to note that for the simple case of a system without a detector, tracking can be successful as long as the object leaves and enters the field-of-view at the same position. A video demo of the tracker-alone setup can be accessed here\footnote{https://youtu.be/6gAMFKbVwAI} for a rigid translational motion case, where the tracker freezes the old location until events occur for continuing the tracking process. Naturally, a tracker-alone setup will fail when the object re-enters at a different position and the proposed eLOT system needs to be employed for the general motion case. A video demo of the eLOT system can be found here\footnote{https://youtu.be/Va9SBX08XQQ}.
\par
Table.~\ref{tab:shapetrack} shows the OS, CLE and IoU of the eLOT tracking system on all three motion cases of the shapes data. The proposed eLOT system achieves an average OS score of 0.6242 on the translational shapes data, which roughly corresponds to being able to track the required shapes for 62.4\% of the time. The sub-pixel CLE scores indicate that during successful overlap with the ground truth, the tracked events have a very close match with the centroid of the ground truth and an IoU score above 0.5 indicate that true positives outweigh the false positives and negatives significantly. The same trend can be observed for the rotational and 6-DOF motion case with predictably lower OS and IoU score compared to the simpler translational motion case. 
\par
For all three motion cases, we found tracking of objects that frequently enter and exit the field-of-view was not a performance limiting factor. For instance, the oval shape in the 6DOF case exits and re-enters fifty times within the minute-long recording, but performance-wise it is as good as the triangle which re-enters only twenty times. The main performance bottleneck was that similar looking objects that appeared later in the event stream, not present in the training phase, was confused with the tracked object frequently. The detector and tracker were both confident of the wrong object in these cases. Future work addressing online training of the detector and online discriminative training of the tracker can potentially address these issues.
\par
Also, in the case of persistent background events surrounding the ROI or amidst clutter, the eLOT tracking system is bound to exhibit failure, as non-maximal suppression would not help in limiting the bounding box size. A sliding window approach is better suited for a complicated scene and this will be undertaken as a future work.

\begin{figure*}[t]
\centering
\subfloat[Shapes data of temporally close sequences.]{\includegraphics[width=3in]{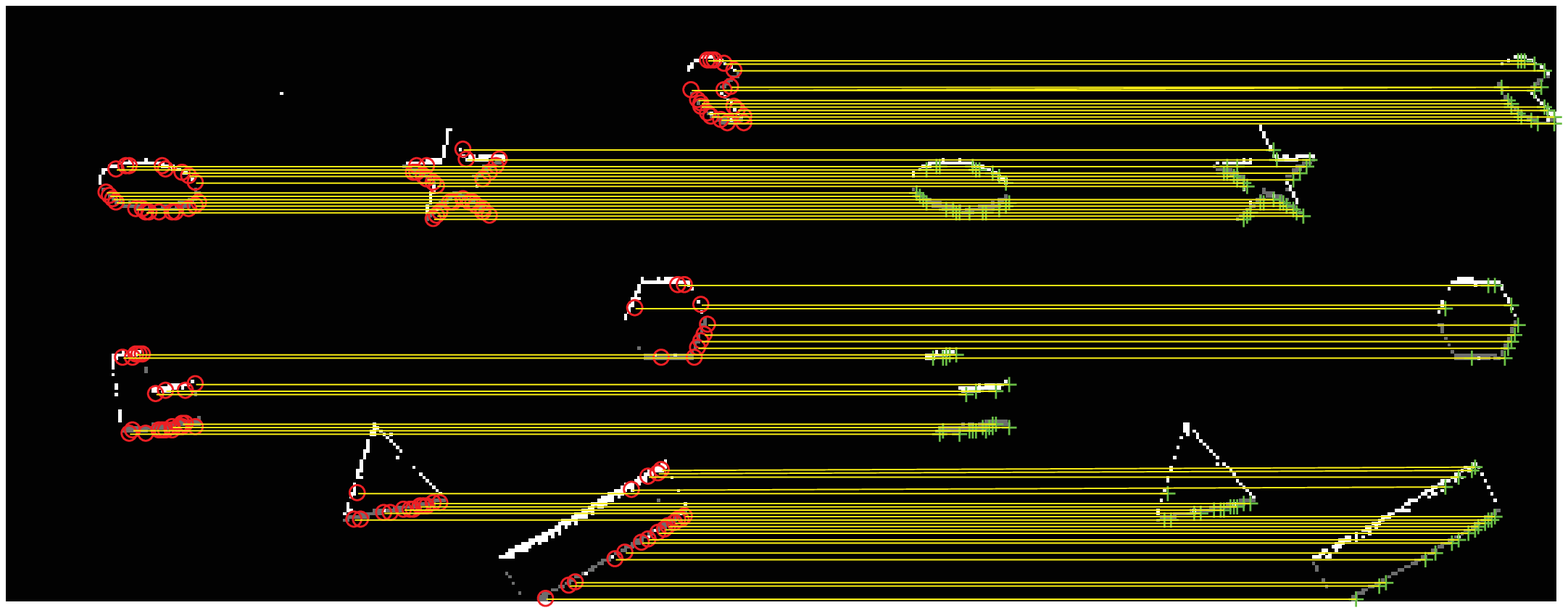}%
\label{fig_shapes_case}}
\hspace{1em}
\subfloat[Shapes data of temporally far sequences with occlusion.]{\includegraphics[width=2.2in]{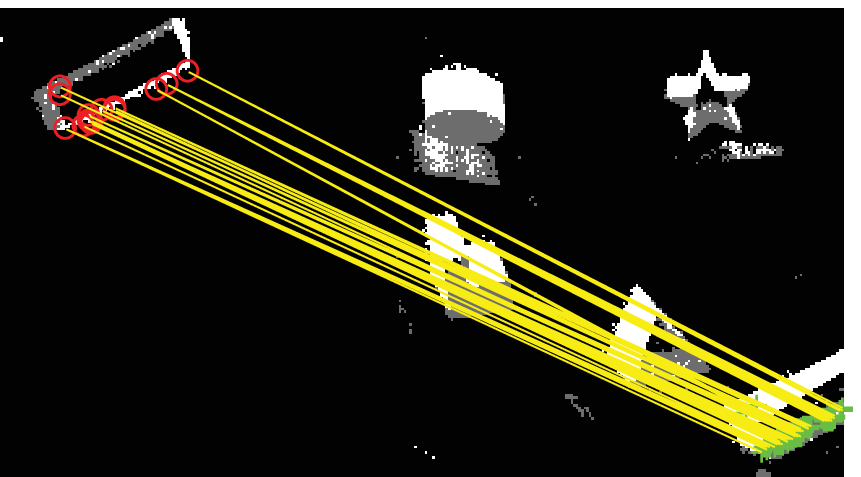}%
\label{fig_rect_case2}}
\hspace{1em}
\subfloat[Indoor scene with different camera motion speeds.]{\includegraphics[width=3.5in]{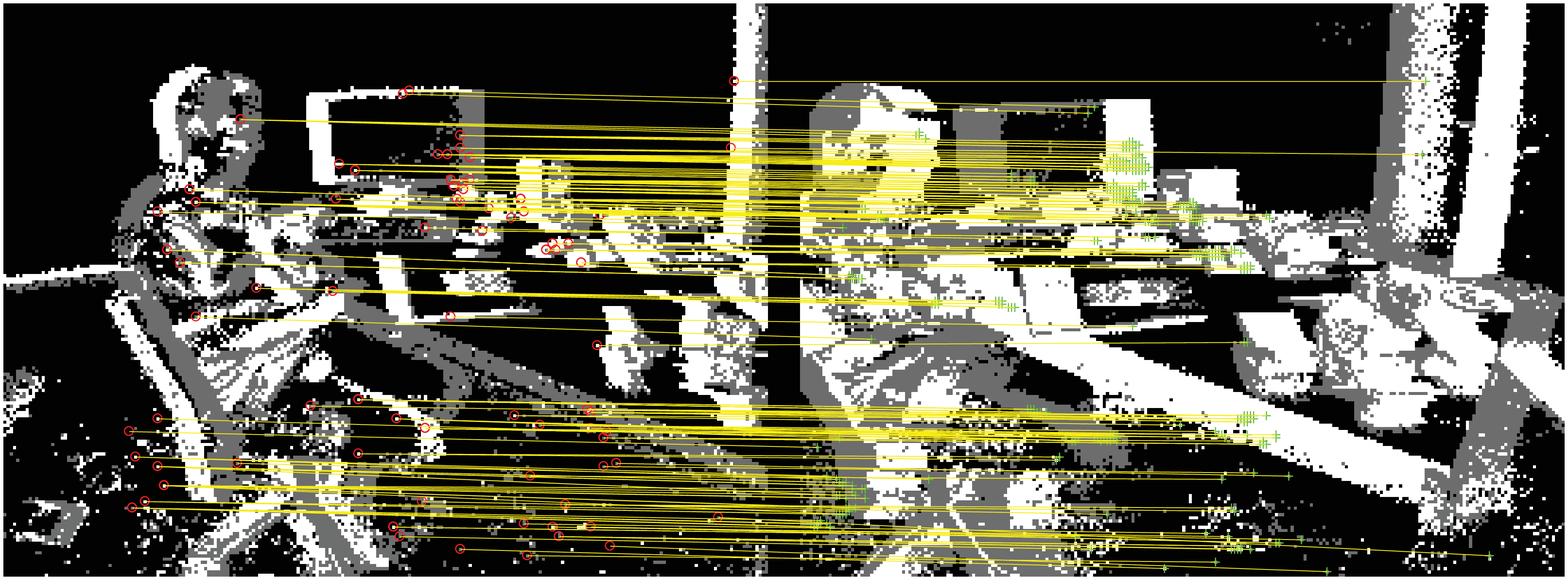}%
\label{fig_simple_case}}
\hspace{1em}
\subfloat[Indoor scene with scale and rotation changes.]{\includegraphics[width=3.5in]{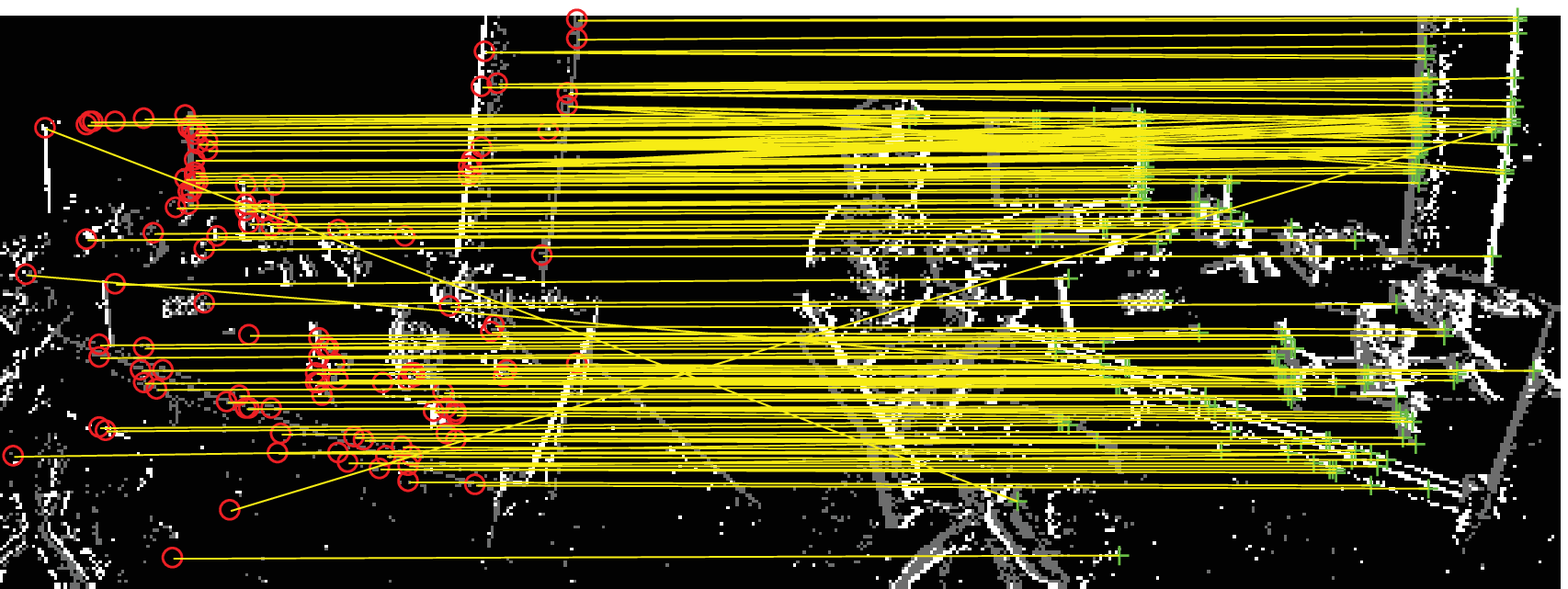}%
\label{fig_complex_case}}
\caption{Feature matching using DART. Top to bottom shows three different scenes. Each scene has two different time-slices of an event camera output displayed side by side. Red circles indicate the features in the left time-slice and the green crosses indicate the corresponding match in the right time-slice connected by yellow lines.}
\label{fig:featmatch}
\end{figure*}

\subsection{Feature Matching}
\label{sec:feature_matching}
Although event cameras output temporally and spatially smooth trajectories, even under fast camera motions, matching events in real-time is still a useful step for vision tasks such as egomotion estimation. For the DART features extracted from two sets of TD events, close matches are found and only a small fraction are incorrect (Fig.~\ref{fig_shapes_case}). As matching events across close time-slices might be sufficient for some tasks, it is also possible to match a single object among several objects present, as shown in Fig.~\ref{fig_rect_case2}. 
\par
For a complex indoor scenes with different camera motion speeds, the robustness of the DART features is shown in Fig.~\ref{fig_simple_case}, where the blur in the second half of the image display indicates a faster camera motion speed (thereby more events) within the same time duration of the time-slice. Finally, Fig.~\ref{fig_complex_case} illustrates feature matching under moderate scale and rotation change of two time-slices far apart in time. 
\section{Conclusion}
\label{sec:9}
This paper presents a novel event-based descriptor, termed as Distribution Aware Retinal Transform (DART), for event cameras that captures precise spatio-temporal structural information using a log-polar grid. Due to the exponential sampling of events, the descriptor is robust to moderate scale and rotation variations. Using the DART descriptor, promising results were demonstrated for four different vision problems, namely object classification, tracking, detection and feature matching. 
\par
For object classification, the DART descriptors were encoded in a bag-of-words framework and testing on several neuromorphic vision datasets was carried out with better results compared to existing works. In addition, multi-scale testing on the MNIST-DVS was carried out to show the tolerance of the proposed descriptors to drastic variations in scale. The classification framework was also demonstrated in real-time using a laboratory setting to show practicality of the proposed approach. In particular, we demonstrated real-time performance on-board a UAV running an Intel Compute Stick which uses an Intel Core m5-6Y57 vPro processor. 
\par
The event-based classification framework was extended to tackle object tracking with two key novelities: (1) training a binary classifier with statistical bootstrapping; (2) robust one-shot learning by applying random circular shifts to the DART descriptors. Additionally, an object detector was designed to solve the event-based long-term object tracking, i.e., re-initialize the tracker after the object exits and enters back the field-of-view. The proposed long-term object tracking system, eLOT, was tested on the event camera dataset by augmenting it object ground truth positions. The annotations are available publicly to create one of the first event-based tracking benchmarks. Finally, we showed the ability of the descriptors to match events across different time-periods and data capturing conditions, which is the basic solution required for many visual tasks such as visual odometry and stereo.


%



\ifCLASSOPTIONcompsoc
  \section*{Acknowledgments}
\else
  \section*{Acknowledgment}
\fi

The authors would like to thank M. S. Muthukaruppan for engaging in extensive discussion and providing valuable comments that have led to substantial improvements of the work.

\ifCLASSOPTIONcaptionsoff
  \newpage
\fi




%

%
%
\begin{IEEEbiography}[{\includegraphics[width=1in,height=1.25in,clip,keepaspectratio]{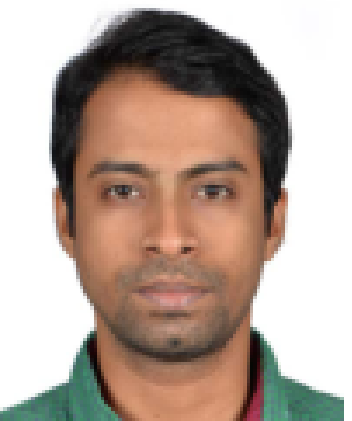}}]{Bharath Ramesh}
 received the B.E. degree in electrical \& electronics engineering from Anna University of India in 2009; M.Sc. and Ph.D. degrees in electrical engineering from National University of Singapore in 2011 and 2015 respectively, working at the Control and Simulation Laboratory on Image Classification using Invariant Features. Bharath’s main research interests include pattern recognition and computer vision. At present, his research is centered on event-based cameras for autonomous robot navigation.
\end{IEEEbiography}
\begin{IEEEbiography}[{\includegraphics[width=1in,height=1.25in,clip,keepaspectratio]{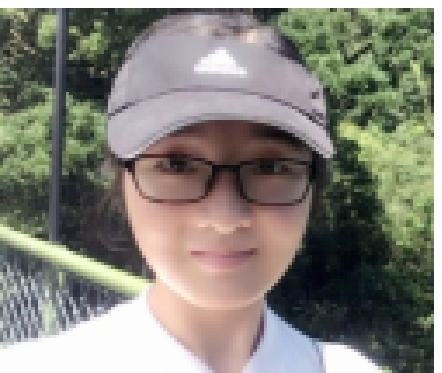}}]{Hong Yang}
 received her Bachelor's degree at University of Electronic Science and Technology of China (UESTC). She is now a master student of NUS and under a working scheme in Temasek Lab. Her current research is on event-based cameras, dealing with real-time pattern recognition problems.
\end{IEEEbiography}
\begin{IEEEbiography}[{\includegraphics[width=1in,height=1.25in,clip,keepaspectratio]{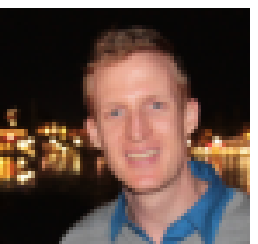}}]{Garrick Orchard}
 holds a B.Sc. degree (with honors, 2006) in electrical engineering from the University of Cape Town, South Africa and M.S.E. (2009) and Ph.D. (2012) degrees in electrical and computer engineering from Johns Hopkins University, Baltimore, USA. His research focuses on developing neuromorphic vision algorithms and systems for real-time sensing on mobile platforms. His other research interests include mixed-signal very large scale integration (VLSI) design, compressive sensing, spiking neural networks, visual perception, and legged locomotion.
\end{IEEEbiography}
\begin{IEEEbiography}[{\includegraphics[width=1in,height=1.25in,clip,keepaspectratio]{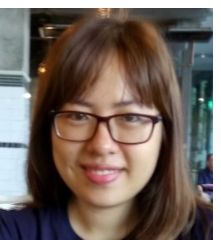}}]{Ngoc~Anh~Le~Thi}
  received her Bachelor's degree at from National University of Singapore in 2017. She finished her Bachelor's thesis on event-based object recognition under the guidance of Dr. Ramesh Bharath, Dr. Garrick Orchard and Assoc Prof Xiang Cheng. She is now a product engineer at Micron Technology, working on NVM bench program development.
\end{IEEEbiography}
\begin{IEEEbiography}[{\includegraphics[width=1in,height=1.25in,clip,keepaspectratio]{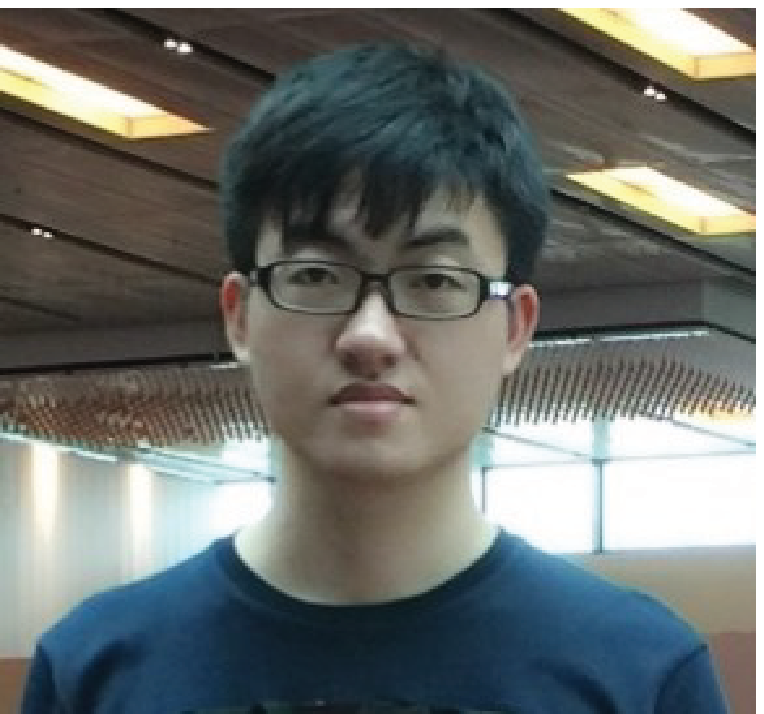}}]{Shihao Zhang}
 is currently an undergraduate at National University of Singapore, studying under the double degree program of computer engineering and economics. He is also under a research intern in Temasek Lab, with focus on event-based object tracking and dealing with real-time problems concerning event-based visual odometry.
\end{IEEEbiography}
\begin{IEEEbiography}[{\includegraphics[width=1in,height=1.25in,clip,keepaspectratio]{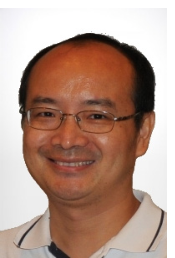}}]{Cheng Xiang}
 received the B.S. degree in mechanical engineering from Fudan University, China in 1991; M.S. degree in mechanical engineering from the Institute of Mechanics, Chinese Academy of Sciences in 1994; and M.S. and Ph.D. degrees in electrical engineering from Yale University in 1995 and 2000, respectively. He is an Associate Professor in the Department of Electrical and Computer Engineering at the National University of Singapore. His research interests include computational intelligence, adaptive systems and pattern recognition.
\end{IEEEbiography}






\end{document}